\documentclass[preprint,3p,12pt,authoryear]{elsarticle}

\usepackage{color}
\usepackage{amssymb, amsfonts, amsthm, amsmath}
\usepackage{graphicx, epstopdf}
\usepackage{cases}
\usepackage{hyperref}
\usepackage{cleveref}
\usepackage{tikz}
\usepackage[utf8]{inputenc}
\usepackage{bbm}
\usepackage{caption}
\usepackage{subcaption}
\usepackage{bm}
\usepackage{empheq}
\usepackage{multirow}
\usepackage{mathtools}
\usepackage{natbib}

\theoremstyle{plain}
\newtheorem{theorem}{Theorem}[section]

\newtheorem{lemma}[theorem]{Lemma}
\newtheorem{assumption}[theorem]{Assumption}
\newtheorem{remark}[theorem]{Remark}

\crefname{assumption}{assumption}{assumptions}
\Crefname{assumption}{Assumption}{Assumptions}

\numberwithin{equation}{section}


\AddToHook{env/subequations/before}{%
}
\AddToHook{env/subequations/after}{%
}

\usepackage{algorithm}
\usepackage{algorithmic}

\journal{}

\begin{document}

\begin{frontmatter}
	
	
	
	\title{FedADMM-InSa: An Inexact and Self-Adaptive ADMM for Federated Learning}
	\nonumnote{Published in \textit{Neural Networks}, volume 181, article 106772, 2025. DOI: \href{https://doi.org/10.1016/j.neunet.2024.106772}{10.1016/j.neunet.2024.106772}.}
	
	\author[label1]{Yongcun Song}
	\ead{ysong307@gmail.com}
	\author[label4]{Ziqi Wang\corref{cor}}
	\cortext[cor]{Corresponding author. Authors are listed alphabetically by family name.}
	\ead{ziqi.wang@fau.de}
	\author[label4,label2,label3]{Enrique Zuazua}
	\ead{enrique.zuazua@fau.de}

	\affiliation[label1]{
		organization={Department of Mathematics, City University of Hong Kong},
		addressline={Kowloon}, 
		city={Hong Kong},
		country={China}}

	\affiliation[label4]{
		organization={Chair for Dynamics, Control, Machine Learning and Numerics – Alexander von Humboldt Professorship, Department of Mathematics, Friedrich-Alexander-Universität Erlangen-Nürnberg},
		addressline={Cauerstrasse 11}, 
		city={Erlangen},
		postcode={91058}, 
		country={Germany}}
	
	\affiliation[label2]{
		organization={Chair of Computational Mathematics, Fundación Deusto},
		addressline={Avenida de las Universidades, 24}, 
		city={Bilbao},
		postcode={48007}, 
		state={Basque Country},
		country={Spain}}
	
	\affiliation[label3]{
		organization={Departamento de Matemáticas, Universidad Autónoma de Madrid},
		addressline={Ciudad Universitaria de Cantoblanco}, 
		city={Madrid},
		postcode={28049}, 
		country={Spain}}
	
	\begin{abstract}
		Federated learning (FL) is a promising framework for learning from distributed data while maintaining privacy. The development of efficient FL algorithms encounters various challenges, including heterogeneous data and systems, limited communication capacities, and constrained local computational resources. Recently developed FedADMM methods show great resilience to both data and system heterogeneity. However, they still suffer from performance deterioration if the hyperparameters are not carefully tuned. To address this issue, we propose an inexact and self-adaptive FedADMM algorithm, termed FedADMM-InSa. First, we design an inexactness criterion for the clients' local updates to eliminate the need for empirically setting the local training accuracy. This inexactness criterion can be assessed by each client independently based on its unique condition, thereby reducing the local computational cost and mitigating the undesirable straggle effect. The convergence of the resulting inexact ADMM is proved under the assumption of strongly convex loss functions. Additionally, we present a self-adaptive scheme that dynamically adjusts each client's penalty parameter, enhancing algorithm robustness by mitigating the need for empirical penalty parameter choices for each client. Extensive numerical experiments on both synthetic and real-world datasets have been conducted. As validated by some tests, our FedADMM-InSa algorithm improves model accuracy by 7.8\% while reducing clients' local workloads by 55.7\% compared to benchmark algorithms.
	\end{abstract}
	
	
	
	
\end{frontmatter}


\section{Introduction}
With the increasing volume of data generated by massively distributed devices and organizations, traditional centralized deep learning paradigms encounter challenges in terms of data collection, privacy concerns, and scalability. To address the above challenges, federated learning (FL)  \citep{li2021survey, li2020federated, mcmahan2017communicationefficient} has emerged as a promising paradigm and has gained significant attention in recent years.   

\subsection{{Background}}
In FL, multiple clients (devices) in a distributed environment collaboratively train a neural network model under the coordination of a central server, without centralizing their data. The training process of FL consists of several communication rounds. For example, in the most commonly used FedAvg \citep{mcmahan2017communicationefficient} algorithm, at every communication round, each client computes a model update on its local data using the latest copy of the global model parameters and then sends the model update to the central server. The server aggregates these updates by averaging to update the global model parameters, which are then sent to all clients. Despite the fact that no raw data is transmitted, there is still a risk of privacy breaches, such as data reconstruction attacks \citep{geng2023improved, wang2023approximatea, xiao2024privacypreserving, yang2023reveal}.

One of the key properties of FL is that the clients are massively distributed and have limited communication capacity \citep{kairouz2021advancesa, mcmahan2017communicationefficient, zhang2022federated}.
To decrease the number of communication rounds needed to train a model, FedAvg suggests increasing local computation on each client. Unlike traditional distributed optimization \citep{zinkevich2010parallelized} where consensus is performed after every local gradient computation, in FedAvg, the clients perform multiple epochs using full-batch or stochastic gradient descents before these local model parameters are aggregated in order to update the global model parameters. 

\subsection{{Current Challenges}}
\textcolor{black}{Nevertheless, simply increasing local training epochs as} in FedAvg would cause the client drift \citep{karimireddy2020scaffold} issue due to the inherent data and system heterogeneity in FL \citep{ kairouz2021advancesa, mcmahan2017communicationefficient}. The client drift issue comes from the fact that clients in FL normally possess unbalanced and non-independent and identically distributed (non-IID) local datasets, which leads to inconsistent minima of the local and global objective functions \citep{karimireddy2020scaffold, wang2020tackling}. As a result, the aggregated global model suffers from deviation and does not converge to its true optimum \citep{li2019convergence}. To address this, FedProx \citep{li2020federateda} was proposed by adding an extra proximal term to the client's loss function to constrain the client's local model to be close to the server's global model. However, the extra proximal term might degrade the training performance unless carefully tuned \citep{wang2022fedadmm}. In \citep{karimireddy2020scaffold}, a method called Scaffold is proposed to reduce client drift by using variance reduction to correct each client's local updates. However, this method doubles the size of variables that need to be communicated between the server and clients, which is not ideal given the limited communication resources in FL.

Recent findings show that FedADMM methods \citep{acar2021federated, gong2022fedadmm,  wang2023admm,zhang2021fedpd, zhou2023federated} based on the alternating direction method of multipliers (ADMM) \citep{boyd2011distributed, glowinski1975approximation} are inherently resilient to heterogeneous clients in FL. FedADMM leverages dual variables to tackle statistical heterogeneity and accommodates system heterogeneity by tolerating variable amounts of work performed by clients. FedADMM maintains identical communication costs per round as FedAvg/Prox and generalizes them using the augmented Lagrangian with dual variables and an extra quadratic penalty term as the client's local loss function. This modification strikes a balance between updating the client's local model and staying consistent with the global model. 

\subsection{{Research Motivations}}
Despite the above-mentioned nice features, current FedADMM methods still suffer from performance deterioration and the straggler effect \citep{kairouz2021advancesa, tan2023personalized} if the hyperparameters are not carefully tuned, especially the amount of local training workload and the choice of the penalty parameter.
In particular, FedADMM methods require clients to perform a certain amount of local training workload, which corresponds to solving the ADMM subproblem inexactly. This is implemented empirically by either commanding clients to solve the subproblem to a constant given accuracy \citep{gong2022fedadmm, zhang2021fedpd, zhou2023federated} or performing a fixed number of local epochs \citep{acar2021federated, wang2023admm}. However, the accuracy of this solution significantly impacts the effectiveness of the algorithm \citep{glowinski2022application}. Meanwhile, empirically assigning the same amount of local training workload overlooks the heterogeneity in clients' data and systems (e.g., non-IID datasets and varying computational resources). This oversight may also cause a severe straggler effect, as waiting for the resource-constrained clients to finish the overload work will slow down the training process, and simply dropping them will lead to a biased global model \citep{bonawitz2019federated, zhou2023federated}.

Another critical issue involves the selection of the penalty parameter of the quadratic term in the augmented Lagrangian functions. It has been shown in ADMM applications \citep{he2000alternating, song2016fast, xu2017admm} that the efficiency of ADMM heavily depends on the penalty parameter. If the penalty parameter is chosen too small or too large, the solution time can increase significantly. The problem is more complicated in FedADMM as each client can use a different penalty parameter, and an inappropriate choice of the penalty parameter can potentially deteriorate the performance of FedADMM methods. 

\subsection{{Main Contributions}}
To address the aforementioned issues, we propose an inexact and self-adaptive FedADMM algorithm, referred to as FedADMM-InSa. Firstly, we introduce an easy-to-implement and adaptive inexactness criterion to guide the client's local training. Our approach eliminates the need to manually set local epochs or predefine a constant accuracy. {\color{black} This provides each client with the flexibility to solve its subproblem inexactly based on its unique situation, thereby eliminating the potential straggler effect.} Furthermore, we design a scheme to dynamically adjust each client's penalty parameter based on the discrepancy between its local model parameters and the global model parameters, avoiding unexpected performance deterioration due to improperly chosen fixed penalty parameters. 

Overall, our main contributions are as follows:
\begin{enumerate}
	\item We propose an inexactness criterion that enables each client to dynamically adjust the precision of local training in each communication round. There is no need to empirically set the number of local training epochs or perceived constant accuracy \textit{a priori}. This flexibility allows our algorithm to better adapt to the heterogeneous clients and datasets, save local computational resources, and mitigate the straggler effect.
	
	\item We develop a self-adaptive scheme for adjusting each client's penalty parameter. The scheme dynamically balances the primal and dual residuals defined by the dissimilarity between the client's local parameters and the server's global parameters between two communication rounds. This adaptive scheme significantly enhances the robustness of our algorithm and eliminates the risk associated with selecting inappropriate penalty parameters for individual clients.
	
	\item The convergence of our proposed algorithm using the inexactness criterion is analyzed, and an $\mathcal{O}(1/K)$ convergence-rate estimate is further established in the ergodic and best-iterate senses. Extensive numerical experiments demonstrate the improved performance of our proposed inexactness criterion and self-adaptive penalty adjusting scheme. As validated by some numerical tests, our proposed algorithm reduces the clients' local computational load by 55.7\% while accelerating the learning process when compared to the vanilla FedADMM.
\end{enumerate}

\subsection{{Organization}}
The rest of the paper is organized as follows. In \Cref{sec_preliminaries}, we first provide a brief background on FL and ADMM, then the vanilla application of ADMM in FL and the resulting FedADMM algorithm. \Cref{sec_our_algorithm_design} presents our proposed FedADMM-InSa method, including the inexactness criterion and the self-adaptive penalty parameter scheme. The experimental setup and simulation results are presented in \Cref{sec_simulations}. Finally, we conclude the paper in \Cref{sec_conclusions} and provide more technical details in Appendix \ref{sec_appendix}.

\section{Preliminaries}\label{sec_preliminaries}
In this section, we briefly introduce the mathematical formulation of FL and ADMM, which are central subjects of the study in this paper.
\subsection{Federated Learning}
Mathematically, the training process of horizontal FL can be formulated as the following minimization problem:
\begin{equation}\label{prob_FL_original}
	\min_{z \in \mathbb{R}^n} \sum_{i=1}^m \alpha_i f_i(z),
\end{equation}
where $m$ is the number of clients, $z \in \mathbb{R}^n$ is the trainable parameters, $f_i: \mathbb{R}^{n} \rightarrow \mathbb{R}$ is the local loss function of each client $i \in [m] \coloneqq  \{1, \ldots, m\}$, $\alpha_i > 0$ is the weight coefficient assigned to client $i$ by the server, and $\sum_{i=1}^m \alpha_i=1$. A common choice is  $\alpha_i=N_i/N$, where $N_i$ is client $i$'s data volume and $N=\sum_{i=1}^m N_i$, i.e., weighting clients proportionally to their data volumes.

\subsection{ADMM}
Given the following constrained optimization problem:
\begin{equation}\label{prob_admm_original}
	\min _{x \in \mathbb{R}^{n_1}, y \in \mathbb{R}^{n_2}} \theta_1(x)+\theta_2(y) \text {, s.t. } A x+B y=b,
\end{equation}
where $\theta_1: \mathbb{R}^{n_1} \rightarrow \mathbb{R}, \theta_2: \mathbb{R}^{n_2} \rightarrow \mathbb{R}$, $A \in \mathbb{R}^{{n_3} \times n_1}, B \in \mathbb{R}^{{n_3} \times n_2}$, and $b \in \mathbb{R}^{n_3}$, its augmented Lagrangian function $L_{\beta}: \mathbb{R}^{n_1} \times \mathbb{R}^{n_2} \times \mathbb{R}^{n_3} \rightarrow \mathbb{R}$ is defined as
\begin{equation}\label{AL_classic_admm}
	L_{\beta}(x, y, \lambda)
	=\theta_1(x)+\theta_2(y)
	-\lambda^T(Ax+By-b) 
	+\frac{\beta}{2}\|Ax+By-b\|^2,
\end{equation}
where $\lambda \in \mathbb{R}^{n_3}$ is the Lagrange multiplier associated with the equality constraint in \eqref{prob_admm_original}, and $\beta>0$ is a penalty parameter. Here and henceforth, we denote by $\|\cdot\|$ the Euclidean (or $\ell_2$-) norm.

Then, starting with an initial point $\{y^0, \lambda^0\}$, for $k\geq 0$, the ADMM \citep{glowinski1975approximation} iteratively updates the variables $\left\{x, y, \lambda\right\}$ in the following way:
\begin{subequations}\label{admm}
	\begin{empheq}[left=\empheqlbrace]{align}
		&x^{k+1}=\arg\min_{x\in \mathbb{R}^{n_1}} L_{\beta}\left(x, y^k, \lambda^k\right), \\
		&y^{k+1}=\arg\min_{y\in \mathbb{R}^{n_2}} L_{\beta}\left(x^{k+1}, y, \lambda^k\right), \\
		&\lambda^{k+1}=\lambda^k-\beta\left( Ax^{k+1} +B y^{k+1}-b \right).
	\end{empheq}
\end{subequations}
{\color{black} The ADMM is a variant of the classic augmented Lagrangian method (ALM) \citep{hestenes1969multiplier, powell1969method}, where the subproblem of each ALM iteration is decomposed into two parts and then solved in a Gauss-Seidel manner. The ADMM possesses a distinct advantage as the decomposed subproblems are typically much easier than those encountered in the ALM, allowing for the exploitation of inherent properties and structures within the model under investigation. Moreover, the ADMM usually exhibits satisfactory numerical performance without the need for specific initial iterates. These characteristics make the ADMM a benchmark algorithm across various domains, including image processing \citep{he2018class}, statistical learning \citep{goldstein2015adaptive,yue2018implementing}, and optimal control problems \citep{glowinski2020admm,glowinski2022application,song2023admmpinnsa,zhang2017alternating}, etc. For a more comprehensive understanding of the ADMM, one can refer to \citep{boyd2011distributed, glowinski2014alternating}.}

\subsection{Vanilla FedADMM}
In this subsection, we present the application of ADMM in the context of FL, leading to the derivation of the vanilla FedADMM \citep{zhou2023federated}. Subsequently, we systematically analyze the challenges and limitations associated with each iterative step of the vanilla FedADMM.

To apply the ADMM to address the FL problem \eqref{prob_FL_original}, we first introduce auxiliary variables $u_i = z$, $i \in[\mathrm{m}]$, and  rewrite problem \eqref{prob_FL_original} into the following consensus setting:
\begin{equation}\label{prob_FL_consensus}
	\min_{u_i, z \in \mathbb{R}^n}  \sum_{i=1}^m \alpha_i f_i(u_i), \text{ s.t. } u_i = z, \forall i \in[m].
\end{equation}
It is clear that problem \eqref{prob_FL_consensus} is equivalent to \eqref{prob_FL_original} in the sense that their optimal solutions coincide. The formulation of problem \eqref{prob_FL_consensus} naturally fits into the FL setting, where $u_i \in \mathbb{R}^n$ can be interpreted as the local model parameters held by client $i \in [m]$, and $z \in \mathbb{R}^n$ as the global model parameters held by the server. 
Therefore, in the following discussions, we concentrate on solving the optimization problem \eqref{prob_FL_consensus} instead of \eqref{prob_FL_original}, as they are equivalent.

For problem \eqref{prob_FL_consensus}, we define its augmented Lagrangian function as
\begin{equation}\label{AL_sum}
	\begin{aligned}
		L_{\alpha, \beta}(u,\lambda,z) &= \sum_{i=1}^m {\alpha_i} L_{\beta_i}(u_i, \lambda_i, z), \text{ with}\\
		L_{\beta_i}(u_i, \lambda_i, z) &= f_i(u_i) - \lambda_i^{\top}\left(u_i-z\right) + \frac{\beta_i}{2}\left\|u_i-z\right\|^2.
	\end{aligned}
\end{equation}
Here, $\lambda_i \in \mathbb{R}^n$ and $\beta_i>0$ are the Lagrange multiplier and the penalty parameter for client $i \in [m]$, respectively. For simplicity, we also denote
\begin{equation}\label{notation_part1}
	\begin{aligned}
		&u=\left(u_1, \ldots ,u_m\right)^{\top} \in \mathbb{R}^{m n}, \
		\lambda=\left(\lambda_1, \ldots, \lambda_m\right)^{\top} \in \mathbb{R}^{m n},\\
		&\alpha=\left(\alpha_1, \ldots ,\alpha_m\right)^{\top} \in \mathbb{R}^m,\
		\beta=\left(\beta_1, \ldots ,\beta_m\right)^{\top} \in \mathbb{R}^m. \\
	\end{aligned}
\end{equation}
Then, given an initial point $\{z^0, \lambda^0\}$, for $k\geq 0$, we can have the vanilla FedADMM that iteratively updates the variables $\left\{u, \lambda, z\right\}$ in the following steps:
\begin{subequations}\label{fedadmm_vanilla}
	\begin{empheq}[left=\empheqlbrace]{align}
		u_i^{k+1} &= \arg\min _{u_i\in \mathbb{R}^n}  L_{\beta_i} \left(u_i, \lambda_i^k, z^k\right), \forall i \in[m],  \label{fedadmm_vanilla_u}
		\\
		\lambda_i^{k+1} &= \lambda_i^k - \beta_i\left( u_i^{k+1} - z^{k} \right), \forall i \in[m], \label{fedadmm_vanilla_l}
		\\ 
		z^{k+1} &= \arg\min_{z \in \mathbb{R}^n}  L_{\alpha, \beta}\left(u^{k+1}, \lambda^{k+1}, z\right). \label{fedadmm_vanilla_z}
	\end{empheq}
\end{subequations}
As seen above, these iterative updates naturally suit the FL setting. The updates of $u_i^{k+1}$ in \eqref{fedadmm_vanilla_u} and $\lambda_i^{k+1}$ in \eqref{fedadmm_vanilla_l} represent client $i$'s local update process, and can be performed in parallel by each client $i \in [m]$. Meanwhile, the update of $z^{k+1}$ in \eqref{fedadmm_vanilla_z} is the server aggregation process after receiving $u^{k+1}$, $\lambda^{k+1}$, and $\beta$ from the clients. We then take a closer look at the solutions of the subproblems \eqref{fedadmm_vanilla_u} - \eqref{fedadmm_vanilla_z} and analyze the associated difficulties and limitations. 

\subsubsection{Client's Local Update: Subproblems \eqref{fedadmm_vanilla_u} and \eqref{fedadmm_vanilla_l}}\label{subsec_client}
The $u$-subproblem \eqref{fedadmm_vanilla_u} and the $\lambda$-subproblem \eqref{fedadmm_vanilla_l} correspond to the client's local update process and can be performed in parallel. The $\lambda$-subproblem \eqref{fedadmm_vanilla_l} is simple and has an analytical form. However, the $u$-subproblem \eqref{fedadmm_vanilla_u} is challenging due to the use of high-dimensional and nonlinear neural networks (such as the ResNets \citep{he2016deep}) in FL. It generally lacks a closed-form solution. As a result, the subproblem \eqref{fedadmm_vanilla_u} should be solved iteratively and inexactly, and the implementation of the FedADMM \eqref{fedadmm_vanilla} must be embedded by an internal iterative process for the subproblem \eqref{fedadmm_vanilla_u}

In the vanilla FedADMM method, the clients normally use gradient-based methods for solving \eqref{fedadmm_vanilla_u} to a certain precision. \textcolor{black}{For example, 
client $i$ first sets $\hat u_i^{k} \coloneqq  z^k$, then implements a prescribed number of local epochs consisting of multiple gradient descent steps:
\begin{equation}\label{fedadmm_vanilla_u_FGD}
	\hat u_i^{k} \coloneqq  \hat u_i^{k} - \eta_i \nabla_{u_i} L_{\beta_i} (\hat u_i^{k}, \lambda_i^k, z^k),
\end{equation}
where $\eta_i>0$ is the learning rate (also known as the step size). 
Then, client $i$ sets $u_i^{k+1} \coloneqq  \hat u_i^{k}$ }and uses it to update $\lambda_i^{k+1}$ by \eqref{fedadmm_vanilla_l}. 
While other optimization schemes like stochastic gradient descent and quasi-Newton methods such as L-BFGS \citep{liu1989limited} are plausible, the accuracy of this inexact solution significantly influences the algorithm's effectiveness \citep{glowinski2022application}. A notable mathematical problem arises concerning the determination of an appropriate inexactness criterion for solving the subproblem \eqref{fedadmm_vanilla_u} inexactly.

Current FedADMM methods often require an empirically perceived precision in advance. For instance, clients may be instructed to perform a fixed number of training epochs \citep{acar2021federated, wang2023admm} or solve the subproblem up to a constant predefined accuracy \citep{gong2022fedadmm, zhang2021fedpd,zhou2023federated}. 
However, empirically assigning the same amount of local training workload is not a good strategy for heterogeneous clients with non-IID data and varying computational resources. For instance, assigning too much workload may inefficiently utilize local computational resources and cause a severe straggler effect, especially among clients with limited computational resources. Moreover, there is no necessity to pursue excessively accurate solutions for \eqref{fedadmm_vanilla_u}, particularly when the iterates are still far from the solution point. Additionally, current FedADMM methods employ a fixed penalty parameter $\beta$ throughout the training process, without considering the  heterogeneity of clients in FL. If the penalty parameter is inappropriately chosen at the beginning, the method's efficiency and performance would deteriorate without remedy.

To tackle the above-mentioned issues, we design a readily implementable and appropriately accurate inexactness criterion for solving subproblem \eqref{fedadmm_vanilla_u} and a self-adaptive scheme to adjust each client's penalty parameter $\beta_i$ in \Cref{sec_our_algorithm_design}. As a result, an inexact and self-adaptive FedADMM is proposed, enabling clients to dynamically adjust local training precision each round, eliminating the need to preset accuracy. This flexibility enhances the adaptability of our algorithm to diverse clients and datasets, saves local computational resources, and mitigates the straggler effect. In addition, the self-adaptive penalty parameter scheme improves the robustness of our algorithm by eliminating the risk associated with inappropriate pre-selection of clients' penalty parameters.
\subsubsection{Server's Aggregation: Subproblem \eqref{fedadmm_vanilla_z}}\label{subsec_server}
The $z$-subproblem \eqref{fedadmm_vanilla_z} is addressed by the server to update the global parameter $z^{k+1}$. It follows from \eqref{AL_sum} that
\begin{equation}
	L_{\alpha, \beta}\left(u^{k+1}, \lambda^{k+1}, z\right) 
	= \sum_{i=1}^m {\alpha_i} \left(f_i(u_i^{k+1}) - (\lambda_i^{k+1})^{\top}(u_i^{k+1}-z) + \frac{\beta_i}{2}\|u_i^{k+1}-z\|^2\right),
\end{equation}
which implies that
\begin{equation}
	z^{k+1} = \arg \min_{z \in \mathbb{R}^n}  \sum_{i=1}^m {\alpha_i} \left((\lambda_i^{k+1})^{\top} z + \frac{\beta_i}{2}\|u_i^{k+1}-z\|^2\right).
\end{equation}
Hence, $z^{k+1}$ has an analytical form given by
\begin{equation}\label{fedadmm_vanilla_z_solution}
	z^{k+1}
	= \frac{1}{\sum_{i=1}^m {\alpha_i} \beta_i} \sum_{i=1}^m {\alpha_i} \left( \beta_i u_i^{k+1} - \lambda_i^{k+1}\right).
\end{equation}
The server can also incorporate the partial client participation strategy by selecting a subset of clients $\mathcal{M}^k \subseteq [m]$ to calculate updated parameters $u_i^{k+1}$ and $\lambda_i^{k+1}$ using \eqref{fedadmm_vanilla_u} and \eqref{fedadmm_vanilla_l} at the $(k+1)$-th communication round. Note from \eqref{fedadmm_vanilla_z_solution} that the clients can provide the server with calculated $(\beta_i u_i^{k+1} - \lambda_i^{k+1})$ instead of $u_i^{k+1}$ and $\lambda_i^{k+1}$, thereby reducing the communication cost. Meanwhile, for the unselected clients $i \notin \mathcal{M}^k$, the server can simply use their previously communicated local model parameters by setting $u_i^{k+1} \coloneqq u_i^{k}$ and $\lambda_i^{k+1} \coloneqq \lambda_i^{k}$. Finally, after receiving updated parameters $(\beta_i u_i^{k+1} - \lambda_i^{k+1})$ and $\beta_i$ from the clients, the server can get the updated global model parameters $z^{k+1}$ by \eqref{fedadmm_vanilla_z_solution}.

With the above analysis of the client update and the server aggregation processes, we summarize the vanilla FedADMM algorithm based on \eqref{fedadmm_vanilla} in \Cref{alg_FedADMM_cano}.
\begin{algorithm}[!ht]
	\caption{Vanilla FedADMM.}
	\label{alg_FedADMM_cano}
	\begin{algorithmic}[1]
		\STATE \textbf{Inputs:} Initialize $z^0, \lambda_i^0, u_i^0, \beta_i, \alpha_i, \eta_i, i \in [m]$.
		\FOR{each communication round $k=0,1, \ldots, K-1$}
		\STATE{\underline{\textbf{Server side:}} Select a subset $\mathcal{M}^k \subseteq [m]$ of clients and send them $z^k$.}
		\STATE{\underline{\textbf{Client side:}}} 
		\FOR{each client $i \in \mathcal{M}^k$ in parallel}
		\STATE{Set $\hat u_i^{k} \coloneqq  z^k$.}
		\FOR{each epoch $e=1, \ldots, E$}
		\STATE \textcolor{black}{Update $\hat u_i^{k} \coloneqq  \hat u_i^{k} - \eta_i \nabla_{u_i} L_{\beta_i} (\hat u_i^{k}, \lambda_i^k, z^k)$ for one epoch.}
		\ENDFOR
		\STATE Set $u_i^{k+1} \coloneqq  \hat u_i^{k}$.
		\STATE Update $\lambda_i^{k+1} = \lambda_i^k-\beta_i( u_i^{k+1}- z^{k} )$. 
		\STATE Send $(\beta_i u_i^{k+1} - \lambda_i^{k+1})$ and ${\beta_i}$ to the server.
		\ENDFOR
		\FOR{each client $i \notin \mathcal{M}^k$ in parallel}
		\STATE Set $u_i^{k+1} \coloneqq u_i^{k}$, $\lambda_i^{k+1} \coloneqq \lambda_i^{k}$.
		\ENDFOR
		\STATE{\underline{\textbf{Server side:}}}
		\STATE Update $z^{k+1} = \frac{1}{\sum_{i=1}^m \alpha_i	 \beta_i} \sum_{i=1}^m \alpha_i (\beta_i u_i^{k+1} - \lambda_i^{k+1})$.
		\ENDFOR
	\end{algorithmic}
\end{algorithm}

\section{An Inexact and Self-Adaptive FedADMM}\label{sec_our_algorithm_design}
In this section, we present the design of our inexact and self-adaptive FedADMM algorithm. We first present a refined server update approach with improved stability. Then, we design an inexactness criterion for solving \eqref{fedadmm_vanilla_u} and a self-adaptive penalty parameter scheme to update $\beta$. We further show that the inexactness criterion and the self-adaptive scheme can be combined with each other.

\subsection{Server's Aggregation with Memory}

From \eqref{fedadmm_vanilla_z_solution}, it is evident that in the vanilla FedADMM, the update of the server's global model parameters $z^{k+1}$ depends solely on the local parameters updated by clients in the $(k+1)$-th communication round. Due to the uncertainty arising from heterogeneous local updates, the global model parameters may undergo significant fluctuations. To address this challenge, we draw inspiration from exponential moving average methods used in several fields (see e.g., \citep{awheda2016exponential, cai2021exponential, dinh2020personalized}) and propose an improved method for updating global model parameters $z^{k+1}$ as follows:
\begin{subequations}\label{fedadmm_with_delta_z}
	\begin{empheq}[left=]{align}
		\hat z^{k+1} &= \frac{1}{\sum_{i=1}^m \alpha_i	 \beta_i} \sum_{i=1}^m \alpha_i \left(\beta_i u_i^{k+1} - \lambda_i^{k+1}\right), \label{fedadmm_with_delta_z_hat}
		\\
		z^{k+1} &= \frac{1}{1+\delta} \hat z^{k+1} + \frac{\delta}{1+\delta} z^{k},  \label{fedadmm_with_delta_z_kplus1}
	\end{empheq}
\end{subequations}
where $\delta > 0$ controls the trade-off between stability and responsiveness in the server update. 

The proposed server update strategy \eqref{fedadmm_with_delta_z} improves the stability of aggregation by balancing the current and previous global model parameters. This ensures the robustness of the global model in the face of variable clients' local updates. On the one hand, it allows for the memorization of $z^k$, thereby minimizing the sensitivity to outliers and leading to a more stable global model. On the other hand, the strategy maintains adaptability to the changing $\hat{z}^{k+1}$, ensuring that the global model responds appropriately to evolving local updates.

Meanwhile, the proposed server update strategy \eqref{fedadmm_with_delta_z} can be interpreted by augmenting the vanilla $z$-subproblem \eqref{fedadmm_vanilla_z} with an additional proximal term, given by
\begin{equation}
	z^{k+1} = \arg \min_{z \in \mathbb{R}^n} \ L_{\alpha, \beta}\left(u^{k+1}, \lambda^{k+1}, z\right) + \frac{\delta}{2} \sum_{i=1}^m \alpha_i \beta_i \|z -z^k\|^2.
\end{equation}

\subsection{An Inexact Version of FedADMM}
In this subsection, we propose an inexactness criterion for solving \eqref{fedadmm_vanilla_u} to address the challenges of presetting the client's local update precision.

\paragraph{Inexactness criterion of \eqref{fedadmm_vanilla_u}}
We first start with analyzing the optimality condition of the $u$-subproblem \eqref{fedadmm_vanilla_u}. From \eqref{AL_sum}, we have
\begin{equation}\label{AL_client_i}
	L_{\beta_i} \left(u_i, \lambda_i^k, z^k\right) = f_i(u_i) - (\lambda_i^k)^{\top}(u_i-z^k) + \frac{\beta_i}{2}\|u_i-z^k\|^2,
\end{equation}
and accordingly
\begin{equation}
	\nabla_{u_i} L_{\beta_i} \left(u_i, \lambda_i^k, z^k\right) 
	=\nabla f_i(u_i)-\lambda_i^k+ \beta_i\left(u_i-z^k\right).
\end{equation}
Let $u_i^{k+1}$ be a solution of the subproblem \eqref{fedadmm_vanilla_u}, then, it satisfies
\begin{equation}\label{opt_con_u_kplus1}
	\nabla f_i(u_i^{k+1})-\lambda_i^k+ \beta_i\left(u_i^{k+1}-z^k\right)=0.
\end{equation}
Based on the optimality condition \eqref{opt_con_u_kplus1}, for each client $i \in [m]$, we define $e_i^k(u_i)$ as
\begin{equation}\label{error_i_k}
	e_i^k(u_i)=\nabla f_i(u_i)-\lambda_i^k+ \beta_i\left(u_i-z^k\right).
\end{equation}
It is clear that a solution $u_i^{k+1}$ of the $u$-subproblem \eqref{fedadmm_vanilla_u} at the $(k+1)$-th iteration satisfies $e_i^k(u_i^{k+1})=0$. Hence, we can use $e_i^k(u_i)$ as the residual for the $u$-subproblem. 
With the help of $e_i^k(u_i)$, we propose the following inexactness criterion. For each client $i \in [m]$, at the $(k+1)$-th iteration, it computes $u_i^{k+1}$ such that
\begin{equation}\label{inexact}
	\left\|e_i^k(u_i^{k+1})\right\| \leq \sigma_i\left\|e_i^k(u_i^k)\right\|,
\end{equation}
where $\sigma_i$ is a given constant satisfying
\begin{equation}\label{sigma}
	0<\sigma_i<\frac{\sqrt{2c_i}}{\sqrt{2c_i}+\sqrt{\beta_i}} < 1,
\end{equation}
and $c_i>0$ is a parameter associated with the strong convexity constant of client $i$'s loss function $f_i$, i.e., $(\nabla f_i(x) - \nabla f_i(y))^{\top}(x-y) \geq c_i\|x-y\|^2, \forall x, y \in  \mathbb{R}^{n}$.
Equivalently, the condition  (\ref{sigma}) can be written as
\begin{equation}\label{sigma_equi}
	0<\sigma_i<\frac{\sqrt{2}}{\sqrt{2}+\sqrt{\tilde{\beta}_i}} < 1,
\end{equation}
where $\tilde{\beta}_i={\beta_i} / {c_i}$. 
It is evident that a smaller $\tilde{\beta}_i$ leads to a higher permissible value of $\sigma_i$, indicating that the subproblem can be solved with less precision. 
In practical scenarios, each client can (locally) convexify its local loss function by incorporating a regularization term. Our inexactness criterion can also be applied to the non-convex problems by using a $\tilde{\beta_i} > 0$ in \eqref{sigma_equi}. Subsequent numerical experiments in \Cref{sec_simulations}, covering both strongly convex and non-convex problems, also underscore the feasibility of our inexactness criterion \eqref{inexact}. 

It is noteworthy that our inexactness criterion can be assessed by each client autonomously based on its present model and penalty parameter, and it can be seamlessly executed during iterations. There is no requirement to predefine any empirically perceived constant accuracy. This gives each client the flexibility to solve its subproblem inexactly, aligning with its distinct non-IID data and computational resources. Consequently, this approach effectively mitigates the potential risk of the straggler effect that may arise with resource-constrained clients. Overall, these attributes make the inexactness criterion \eqref{inexact} straightforward to implement and more likely to result in local computational savings.

\paragraph{FedADMM-In: An inexact FedADMM algorithm}
Based on the discussions above, the iterative three steps of FedADMM with the inexactness criterion \eqref{inexact} are given by
\begin{subequations}\label{fedadmm_in}
	\begin{empheq}[left=\empheqlbrace]{align}
		u_i^{k+1}& \approx \arg\min_{u_i \in \mathbb{R}^n} L_{\beta_i} \left(u_i, \lambda_i^k, z^k\right), \forall i \in[m], \label{fedadmm_in_u}
		\\
		\lambda_i^{k+1}&=\lambda_i^k-\beta_i\left( u_i^{k+1}- z^{k} \right),\forall i \in[m], \label{fedadmm_in_l} 
		\\ 
		z^{k+1} &= \arg \min_{z \in \mathbb{R}^n} \ L_{\alpha, \beta}\left(u^{k+1}, \lambda^{k+1}, z\right) + \frac{\delta}{2} \sum_{i=1}^m \alpha_i \beta_i \|z -z^k\|^2.\label{fedadmm_in_z}
	\end{empheq}
\end{subequations}
We denote by FedADMM-In the FedADMM algorithm with the inexactness criterion \eqref{inexact} and summarize it in \Cref{alg_FedADMM_in}. 
\begin{algorithm}[!ht]
	\caption{FedADMM-In: FedADMM with the inexactness criterion \eqref{inexact}.}
	\label{alg_FedADMM_in}
	\begin{algorithmic}[1]
		\STATE \textbf{Inputs:} Initialize $z^0, \lambda_i^0, u_i^0, \beta_i, \sigma_i, \alpha_i, \eta_i, i \in [m]$.
		\FOR{each communication round $k=0,1, \ldots, K-1$}
		\STATE{\underline{\textbf{Server side:}} Select a subset $\mathcal{M}^k \subseteq [m]$ of clients and send them $z^k$.}
		\STATE{\underline{\textbf{Client side:}}} 
		\FOR{each client $i \in \mathcal{M}^k$ in parallel}
		\STATE Set $\hat u_i^{k} \coloneqq  z^k$.
		\WHILE{$\|e_i^k(\hat u_i^{k})\| > \sigma_i \|e_i^k(u_i^{k})\|$}
		\STATE \textcolor{black}{Update $\hat u_i^{k} \coloneqq  \hat u_i^{k} - \eta_i \nabla_{u_i} L_{\beta_i} (\hat u_i^{k}, \lambda_i^k, z^k)$ for one epoch.}
		\ENDWHILE
		\STATE{Set $u_i^{k+1} \coloneqq  \hat u_i^{k}$.}
		\STATE Update $\lambda_i^{k+1} = \lambda_i^k-\beta_i( u_i^{k+1}- z^{k} )$. 
		\STATE{Send $(\beta_i u_i^{k+1} - \lambda_i^{k+1})$ and ${\beta_i}$ to the server.}
		\ENDFOR
		\FOR{each client $i \notin \mathcal{M}^k$ in parallel}
		\STATE{Set $u_i^{k+1} \coloneqq  u_i^{k}$, $\lambda_i^{k+1} \coloneqq  \lambda_i^{k}$.}
		\ENDFOR
		\STATE{\underline{\textbf{Server side:}}}
		\STATE Obtain $\hat z^{k+1} = \frac{1}{\sum_{i=1}^m \alpha_i	 \beta_i} \sum_{i=1}^m \alpha_i (\beta_i u_i^{k+1} - \lambda_i^{k+1})$. 
		\STATE Update $z^{k+1} = \frac{1}{1+\delta} \hat z^{k+1} + \frac{\delta}{1+\delta} z^{k}$.
		\ENDFOR
	\end{algorithmic}
\end{algorithm}

\paragraph{Convergence analysis}
Let us analyze the convergence of our proposed FedADMM-In algorithm under the following assumptions:	
\begin{assumption}\label{ass_lipschitz}
	For all $i \in [m]$, the gradient of $f_i$ is $s_i$-Lipschitz, that is,
	\begin{equation}\label{ass1}
		\left\|\nabla f_i(x)-\nabla f_i(y)\right\| \leq s_i\|x-y\|, s_i>0, \forall x, y \in  \mathbb{R}^{n}.
	\end{equation}
\end{assumption}
\begin{assumption}\label{ass_strongly_convex}
	For all $i \in [m]$, $f_i$ is $c_i$-strongly convex, that is,
	\begin{equation}\label{strongly_convex}
		(\nabla f_i(x) - \nabla f_i(y))^{\top}(x-y) \geq c_i\|x-y\|^2, c_i>0, \forall x, y \in  \mathbb{R}^{n}.
	\end{equation}
\end{assumption}
From the strong convexity assumption, it is easy to deduce that problem \eqref{prob_FL_original} has a unique solution, denoted by $z^*$. As a consequence, problem \eqref{prob_FL_consensus}, which is equivalent to \eqref{prob_FL_original}, also has a unique solution $(u^*,z^*)$, where $u_i^* = z^*, \forall i \in [m]$. On the other hand, from the above assumptions, the dual problem of problem \eqref{prob_FL_consensus} has a unique solution $\lambda^*$ and the strong duality holds. In the following \Cref{thm_fedadmmIn}, we show that the sequence generated by iterative scheme \eqref{fedadmm_in} (i.e., \Cref{alg_FedADMM_in} with full client participation) converges to $(u^*,\lambda^*,z^*)$.

\begin{theorem}\label{thm_fedadmmIn}
	Let $\{w^k\}=\{(u^k,\lambda^k, z^k)^{\top}\}$ be the sequence generated by iterative scheme \eqref{fedadmm_in} (i.e., \Cref{alg_FedADMM_in} with full client participation). Then, we have the following assertions:
	\begin{equation}
		\|e_i^k(u_i^{k+1})\| \stackrel{k \rightarrow \infty}{\longrightarrow} 0, \quad
		u_i^k \stackrel{k \rightarrow \infty}{\longrightarrow} u_i^*, \quad
		\lambda_i^k \stackrel{k \rightarrow \infty}{\longrightarrow} \lambda_i^*, \quad
		z^k \stackrel{k \rightarrow \infty}{\longrightarrow} z^*, \quad \forall i \in [m].
	\end{equation}
\end{theorem}
The detailed proof is presented in Appendix \ref{sec_convergence}, which is inspired by \cite[Sec.\@ 3.3]{glowinski2022application}. Although the convergence proof is conducted under the strong convexity assumption, our algorithm with the proposed inexactness criterion can also be applied to the non-convex cases by using a $\tilde{\beta_i} > 0$ in \eqref{sigma_equi}. Moreover, the effectiveness of the algorithm with partial client participation in both strongly convex and non-convex cases is validated by the numerical experiments in \Cref{sec_simulations}. 

\begin{remark}
	In practical applications, one can also use $z^k$ instead of $u_i^k$ on the right-hand side of the inexactness criterion \eqref{inexact}. We empirically demonstrate in \Cref{sec_simulations} that such a substitution improves the FL training performance in terms of improved model accuracy and reduced computational load. This substitution alleviates the unnecessary pursuit of the client's local minima, which is normally inconsistent with the global minima due to the non-IID datasets. The convergence analysis of the algorithm employing this modified inexactness criterion and partial client participation serves as an interesting future study.
\end{remark}

\subsection{Self-Adaptive Penalty Parameter $\beta_i^k$}
\label{sec_adp}
In this subsection, we present a self-adaptive scheme for adjusting the penalty parameter $\beta_i^k$ for client $i$ at communication round $k$ to further improve the performance of our FedADMM-In algorithm. This also makes the algorithm more robust to different initial choices of the penalty parameter. 

\paragraph{Self-adaptive scheme}
As shown in \eqref{fedadmm_in}, the FedADMM-In algorithm is an iterative algorithm, and an intuitive stopping criterion of the iterative updates can be $u_i^{k+1} - u_i^{k} = 0$ and $\lambda_i^{k+1} - \lambda_i^{k} = 0$. It follows from \eqref{fedadmm_in_l} that $\lambda_i^{k+1} - \lambda_i^{k} = 0$ implies that $u_i^{k+1} - z^{k}=0$. 
Inspired by the stopping criterion, we define the primal residual $p_i^k$ and the dual residual $d_i^k$ as follows:
{\begin{subequations}\label{pri_dua_residual}
		\begin{empheq}[left=]{align}
			p_i^k &= \beta_i^k \|u_i^{k+1} - u_i^{k}\|,\  i \in [m], \label{pri_residual}
			\\
			d_i^k &= \|u_i^{k+1}- z^{k}\|, \ i \in [m]. \label{dua_residual}
		\end{empheq}
\end{subequations}}
To adaptively update $\beta_i^k$, we use the following scheme:
\begin{equation}\label{self_adp_beta_rule}
	\beta_i^{k+1}= 
	\begin{cases}
		\beta_i^k \tau, & \text { if } d_i^k>\mu p_i^k, \\ 
		\beta_i^k / \tau, & \text { if } p_i^k>\mu d_i^k, \\ 
		\beta_i^k, & \text { otherwise },
	\end{cases}
\end{equation}
where $\mu, \tau>1$ are parameters to choose. 

The essence of the self-adaptive penalty parameter scheme \eqref{self_adp_beta_rule} lies in dynamically balancing the primal residual $p_i^k$ and the dual residual $d_i^k$. {For instance, according to the definition of $d_i^k$ in \eqref{dua_residual}, $d_i^k>\mu p_i^k$ suggests that client $i$'s updated local model parameters $u_i^{k+1}$ deviate significantly from the current global model parameters $z^{k}$. Consequently, client $i$ is prompted to increase its penalty parameter $\beta_i^k$ to impose a more substantial penalty on the constraint.} Conversely, a reduction in $\beta_i^k$ is recommended for client $i$ to mitigate the primal residual.

Finally, the iterative steps of FedADMM with an adaptive $\beta_i^k$ are given by
\begin{subequations}\label{fedadmm_insa}
	\begin{empheq}[left=\empheqlbrace]{align}
		u_i^{k+1}& \approx \arg\min_{u_i \in \mathbb{R}^n} L_{\beta_i^k} \left(u_i, \lambda_i^k, z^k\right), \forall i \in[m], \label{fedadmm_insa_u}
		\\ 
		\lambda_i^{k+1}&=\lambda_i^k-\beta_i^k\left( u_i^{k+1}- z^{k} \right),\forall i \in[m], \label{fedadmm_insa_l} 
		\\ 
		z^{k+1} &= \arg \min_{z \in \mathbb{R}^n} \ L_{\alpha, \beta^k}\left(u^{k+1}, \lambda^{k+1}, z\right) + \frac{\delta}{2} \sum_{i=1}^m \alpha_i \beta_i^k \|z -z^k\|^2. \label{fedadmm_insa_z}
	\end{empheq}
\end{subequations}

\paragraph{Compatibility with the inexactness criterion}
We can easily extend our inexactness criterion \eqref{inexact} to make it compatible with the self-adaptive penalty parameter scheme \eqref{self_adp_beta_rule}. To accomplish this, we simply replace the fixed $\beta_i$ with the adaptive $\beta_i^k$ in \eqref{sigma}. Then, for each client $i \in [m]$, at the $(k+1)$-th communication round, it computes $u_i^{k+1}$ such that
\begin{equation}\label{inexact_adp}
	\left\|e_i^k(u_i^{k+1})\right\| \leq \sigma_i^k\left\|e_i^k(u_i^k)\right\|,
\end{equation}
where $\sigma_i^k$ is a given constant satisfying 
\begin{equation}\label{sigma_adp}
	0<\sigma_i^k<\frac{\sqrt{2}}{\sqrt{2}+\sqrt{\tilde{\beta}_i^k}} < 1,
\end{equation}
and $\tilde{\beta}_i^k = {\beta_i^k} / {c_i}$ can be independently evaluated by each client $i$ based on its unique situation. Hence, the inexactness criterion \eqref{inexact_adp} with the self-adaptive penalty parameter scheme can also be executed autonomously by each client during iterations. Moreover, this allows clients to perform personalized local update steps based on their distinct data and computational resources, effectively mitigating the risk of the straggler effect.

\paragraph{FedADMM-InSa: An inexact and self-adaptive FedADMM}
Based on the discussions above, we propose the FedADMM-InSa algorithm using an adaptive penalty parameter $\beta_i^k$ and present it in \Cref{alg_FedADMM_insa}. 
It brings more difficulty to prove the convergence of \Cref{alg_FedADMM_insa} with a varying $\beta_i^k$ in each iteration, however, the numerical experiments in \Cref{sec_simulations} empirically validate its robustness and improved performance.
\begin{algorithm}[!h]
	\caption{FedADMM-InSa: FedADMM with the inexactness criterion \eqref{inexact_adp} and the self-adaptive penalty parameter scheme \eqref{self_adp_beta_rule}.}
	\label{alg_FedADMM_insa}
	\begin{algorithmic}[1]
		\STATE \textbf{Inputs:} Initialize $z^0, \lambda_i^0, u_i^0, \beta_i^0, \sigma_i^0, c_i, \alpha_i, \eta_i, i \in [m]$, $\mu>1, \tau>1$.
		\FOR{each communication round $k=0,1, \ldots, K-1$}
		\STATE{\underline{\textbf{Server side:}} Select a subset $\mathcal{M}^k \subseteq [m]$ of clients and send them $z^k$.}
		\STATE{\underline{\textbf{Client side:}}} 
		\FOR{each client $i \in \mathcal{M}^k$ in parallel}
		\STATE Set $\hat u_i^{k} \coloneqq  z^k$.
		\WHILE{$\|e_i^k(\hat u_i^{k})\| > \sigma_i^{k} \|e_i^k(u_i^{k})\|$}
		\STATE \textcolor{black}{Update $\hat u_i^{k} \coloneqq  \hat u_i^{k} - \eta_i \nabla_{u_i} L_{\beta_i} (\hat u_i^{k}, \lambda_i^k, z^k)$ for one epoch.}
		\ENDWHILE
		\STATE{Set $u_i^{k+1} \coloneqq  \hat u_i^{k}$.}
		\STATE Update $\lambda_i^{k+1} = \lambda_i^k-\beta_i^k( u_i^{k+1}- z^{k} )$. 
		\STATE{Send $(\beta_i^k u_i^{k+1} - \lambda_i^{k+1})$ and ${\beta_i^k}$ to the server.}
		\STATE Calculate $p_i^k = \beta_i^k\|u_i^{k+1} - u_i^{k}\|$ and $d_i^k = \|u_i^{k+1} - z^{k}\|$. 
		\IF{$d_i^k>\mu p_i^k$} 
		\STATE Update $\beta_i^{k+1} = \beta_i^k  \tau$.
		\ELSIF{$p_i^k>\mu d_i^k$}
		\STATE Update $\beta_i^{k+1} = \beta_i^k / \tau$.
		\ELSE
		\STATE Update $\beta_i^{k+1} = \beta_i^k$.
		\ENDIF
		\STATE Update $\sigma_i^{k+1} = \frac{\sqrt{2}}{\sqrt{2}+\sqrt{{\beta_i^{k+1}} / {c_i}}}$. 
		\ENDFOR
		\FOR{each client $i \notin \mathcal{M}^k$ in parallel}
		\STATE Set $u_i^{k+1} \coloneqq  u_i^{k}$, $\lambda_i^{k+1} \coloneqq  \lambda_i^{k}$, $\beta_i^{k+1} \coloneqq  \beta_i^k$, $\sigma_i^{k+1} \coloneqq  \sigma_i^k$. 
		\ENDFOR
		\STATE{\underline{\textbf{Server side:}}}
		\STATE Obtain $\hat z^{k+1} = \frac{1}{\sum_{i=1}^m \alpha_i	 \beta_i^k} \sum_{i=1}^m \alpha_i (\beta_i^k u_i^{k+1} - \lambda_i^{k+1})$.
		\STATE Update $z^{k+1} = \frac{1}{1+\delta} \hat z^{k+1} + \frac{\delta}{1+\delta} z^{k}$. 
		\ENDFOR
	\end{algorithmic}
\end{algorithm}

\section{Experimental Results}\label{sec_simulations}
In this section, we conduct extensive numerical tests to demonstrate the improved performance of our proposed FedADMM-In (\Cref{alg_FedADMM_in}) and FedADMM-InSa (\Cref{alg_FedADMM_insa}) in comparison to the vanilla FedADMM (\citep{zhou2023federated}) and {FedAvg \citep{mcmahan2017communicationefficient}}. We first introduce the experimental settings and the implementation details used in our experiments. Then, we present the simulation results and analysis. The source code is available at {\small\url{https://github.com/DCN-FAU-AvH/FedADMM-InSa}.}

\subsection{Setups}
{We conducted experiments on three combinations of datasets and models, covering both cross-device (many clients with few data points per client) and cross-silo (few clients with many data points per client) scenarios \citep{kairouz2021advancesa}. The details of the experiments are elaborated below and also summarized in \Cref{tab_test_params}.}
\paragraph{Example 1: Linear regression with a synthetic dataset}
In this example, we set each client's local loss function to be
\begin{equation}
	f_i(u_i) = \frac{1}{2N_i}\sum_{j=1}^{N_i} \left(u_i^{\top}a_i^j  -b_i^j\right)^2 +\frac{\gamma_i}{2}\|u_i\|^2, \quad i\in [m].
\end{equation}
Here, $a_i^j \in \mathbb{R}^{n_{a}}$ and $b_i^j \in \mathbb{R}$ are the $j$-th data of client $i\in [m]$, $N_i$ is the volume of data of client $i$, and we denote $N= \sum_{i=1}^{m}N_i$. Let $\lceil \cdot \rceil$ be the ceiling function, i.e., $\lceil x \rceil$ is the smallest integer not smaller than $x$, e.g., $\lceil 1.5\rceil=2$. Then, following the setup in \citep{zhou2023federated}, we generate $\lceil N / 3\rceil$ samples from the standard normal distribution, $\lceil N / 3\rceil$ samples from the Student's $t$ distribution with degree 5, and $N-2\lceil N / 3\rceil$ samples from the uniform distribution in $[-5,5]$. In the tests of Example 1, we set {$n_{a}=5,000$, $N = 50,000$}, and $\gamma_i=0.01$ for all $i \in [m]$.

\paragraph{Example 2: Image classification with the MNIST dataset}
In the second example, we address the image classification problem using convolutional neural networks (CNN) with the MNIST dataset  \citep{lecun1998gradientbased}. The MNIST dataset contains images of handwritten digits. It has 60,000 training data and 10,000 test data. The samples in this dataset are $28 \times 28$ grayscale images with handwritten digits from 0 to 9 in the center. {In this example, each client uses a CNN with two convolutional layers, comprising 32 and 64 channels respectively, as the same to that in \citep{mcmahan2017communicationefficient}.}

\paragraph{Example 3: Image classification with the CIFAR-10 dataset}
	In the third example, we test the image classification of the CIFAR-10 dataset  \citep{krizhevsky2009learning}. This dataset contains color images of size $32 \times 32$ with 10 categories. There are 50,000 training data and 10,000 test data. For the neural networks, each client uses a ResNet-20 \citep{he2016deep}, which consists of 20 stacked weighted layers.

\paragraph{Non-IID data separation}
To test the performance of the algorithms on non-IID data, we separate datasets in the following way. In Example 1, we first shuffle the training samples and then distribute them evenly among all clients. In Example 2 and 3, we first sorted the data by labels and then divided them into several shards, each shard containing images with the same label. Then, we assign the data to make sure that each client has at most two kinds of labels in Example 2 and five kinds of labels in Example 3, corresponding to pathological non-IID scenarios. 

\paragraph{FL training parameters}
In each round, the server uniformly samples 20\% of the clients to perform the local training. For the vanilla FedADMM and FedAvg algorithms, the clients perform fixed epochs of training using stochastic gradient descent. 
\textcolor{black}{For the FedADMM-In and FedADMM-InSa algorithms, the number of epochs is controlled by the inexactness criterion, which is evaluated after each epoch.} We constrain the maximum number of epochs per round to be the same as the fixed epochs in the vanilla FedADMM. For the inexactness criterion, we set $c_i=0.01$ in Example 1 and 2, and $c_i=0.001$ in Example 3. In all examples, we replace $u_i^k$ with $z^k$ and set $\delta = 0.01$ for the server aggregation. For FedADMM-InSa, the adaptive penalty scheme uses $\mu=5$ and $\tau=2$ in all examples. Finally, important parameters used in the tests are summarized in \Cref{tab_test_params}.

\begin{table}[!ht]
	\centering
	\caption{{Datasets, models, and parameters of three examples.}}
	\label{tab_test_params}	
		\begin{tabular}{l|c|c|c}
			\hline
			& \begin{tabular}[c]{@{}c@{}}\textbf{Example 1}\\ Linear regression\end{tabular} & \begin{tabular}[c]{@{}c@{}}\textbf{Example 2}\\ Image classification\end{tabular} & \begin{tabular}[c]{@{}c@{}}\textbf{Example 3}\\ Image classification\end{tabular} \\ \hline
			Dataset                     & Synthetic  & MNIST          & CIFAR-10  \\
			Model      & Linear  & CNN          & ResNet  \\
			Training set size                     & 50000  & 60000          & 50000  \\
			Test set size                         & -      & 10000           & 10000  \\
			Data dimension                        & 5000   & $28 \times 28 \times 1$ & $32 \times 32 \times3$  \\
			Number of clients                 & 200    & 200            & 10  \\
			Data per client                 & 250    & 300            & 5000  \\
			Active clients per round  & 20\%     & 20\%             & 20\%  \\
			Client learning rate         & 0.001   & 0.01           & 0.001  \\
			Batch size              & 50    & 50            & 500  \\
			Epochs per round        & 20    & 20            & 10  \\
			Communication rounds              & 300    & 300            & 300  \\
			\hline
	\end{tabular}
\end{table}

\begin{table}[!ht]
	\centering
	\caption{{Comparison results of different examples. The columns showing epoch reduction use the vanilla FedADMM and FedAvg as the baseline.}}
	\label{tab_results_three_egs}
		\begin{tabular}{cc|cc|cc|cc}
			\hline
			&
			&
			\multicolumn{2}{c|}{\begin{tabular}[c]{@{}c@{}}\textbf{Example 1}\\ Linear regression\end{tabular}} &
			\multicolumn{2}{c|}{\begin{tabular}[c]{@{}c@{}}\textbf{Example 2}\\ MNIST, CNN\end{tabular}} &
			\multicolumn{2}{c}{\begin{tabular}[c]{@{}c@{}}\textbf{Example 3}\\ CIFAR-10, ResNet-20\end{tabular}} \\ \hline
			\multicolumn{1}{c|}{$\beta_i$} &
			Algorithm &
			Loss &
			\begin{tabular}[c]{@{}c@{}}Epoch\\ reduction\end{tabular} &
			Accuracy &
			\begin{tabular}[c]{@{}c@{}}Epoch\\ reduction\end{tabular} &
			Accuracy &
			\begin{tabular}[c]{@{}c@{}}Epoch\\ reduction\end{tabular} \\ \hline
			\multicolumn{1}{c|}{-}                    & Fedavg       & 1.64          & -      & 96.0\%          & -      & 36.3\%          & -      \\ \hline
			\multicolumn{1}{c|}{\multirow{3}{*}{0.1}} & FedADMM      & 1.64          & -      & \textbf{98.6\%} & -      & 51.1\%          & -      \\
			\multicolumn{1}{c|}{}                     & FedADMM-In   & 1.63          & 94.3\% & 98.5\%          & 92.4\% & 41.0\%          & 71.5\% \\
			\multicolumn{1}{c|}{}                     & FedADMM-InSa & \textbf{1.51} & 20.3\% & 97.8\%          & 66.9\% & \textbf{53.1\%} & 14.4\% \\ \hline
			\multicolumn{1}{c|}{\multirow{3}{*}{1}}   & FedADMM      & 1.53          & -      & 97.7\%          & -      & \textbf{52.9\%} & -      \\
			\multicolumn{1}{c|}{}                     & FedADMM-In   & 1.55          & 58.5\% & 97.6\%          & 58.7\% & 51.6\%          & 9.4\%  \\
			\multicolumn{1}{c|}{}                     & FedADMM-InSa & \textbf{1.51} & 18.8\% & \textbf{97.9\%} & 61.0\% & 52.1\%          & 9.4\%  \\ \hline
			\multicolumn{1}{c|}{\multirow{3}{*}{2}}   & FedADMM      & 1.51          & -      & 96.2\%          & -      & 46.7\%          & -      \\
			\multicolumn{1}{c|}{}                     & FedADMM-In   & 1.51          & 19.3\% & 96.7\%          & 41.7\% & 45.7\%          & 2.1\%  \\
			\multicolumn{1}{c|}{}                     & FedADMM-InSa & \textbf{1.51} & 16.2\% & \textbf{97.7\%} & 57.9\% & \textbf{50.4\%} & 6.9\%  \\ \hline
			\multicolumn{1}{c|}{\multirow{3}{*}{5}}   & FedADMM      & 1.51          & -      & 92.5\%          & -      & 39.5\%          & -      \\
			\multicolumn{1}{c|}{}                     & FedADMM-In   & 1.51          & 3.9\%  & 93.4\%          & 31.1\% & 39.3\%          & 0.0\%  \\
			\multicolumn{1}{c|}{}                     & FedADMM-InSa & \textbf{1.51} & 12.5\% & \textbf{97.9\%} & 55.9\% & \textbf{51.5\%} & 4.0\%  \\ \hline
			\multicolumn{1}{c|}{\multirow{3}{*}{10}}  & FedADMM      & 1.51          & -      & 89.7\%          & -      & 34.0\%          & -      \\
			\multicolumn{1}{c|}{}                     & FedADMM-In   & 1.51          & 0.9\%  & 90.2\%          & 21.1\% & 37.3\%          & 0.0\%  \\
			\multicolumn{1}{c|}{}                     & FedADMM-InSa & \textbf{1.51} & 6.8\%  & \textbf{97.5\%} & 55.7\% & \textbf{49.8\%} & 4.8\%  \\ \hline
		\end{tabular}
\end{table}

	\subsection{Results Analysis}
	In this subsection, we present the comparison results of our FedADMM-In and FedADMM-InSa algorithms with benchmark FedADMM and FedAvg algorithms. For ADMM-based algorithms, we show results with different values of $\beta_i \in \{0.1, 1, 2, 5, 10\}$. The detailed results are summarized in \Cref{tab_results_three_egs} and elaborated below.
	
	\subsubsection{Evaluation of FedADMM-In} 
\begin{figure}[!ht]
		\centering
		\begin{subfigure}[b]{0.3\textwidth}
			\centering
			\includegraphics[width=\textwidth]{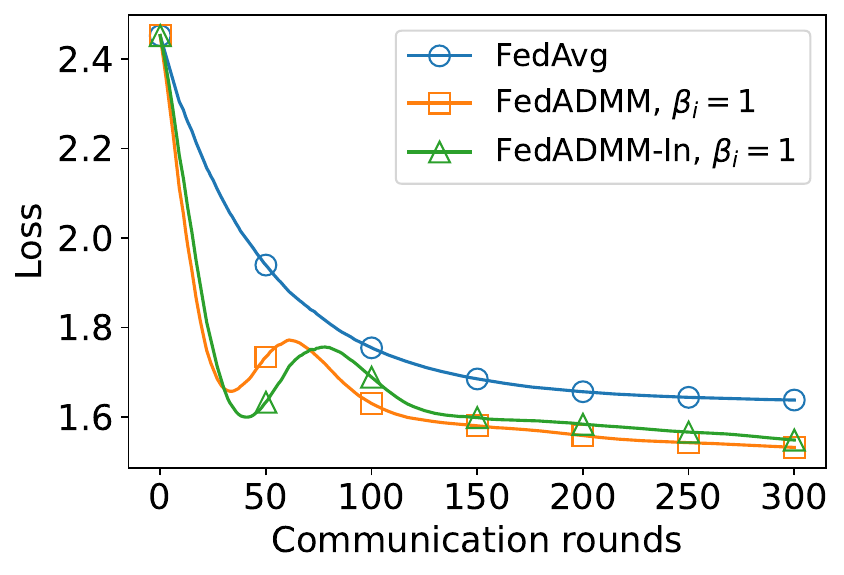}
			\caption{Linear regression.}
			\label{fig_cmp_in_a}
		\end{subfigure}
		\hspace{10pt}
		\begin{subfigure}[b]{0.31\textwidth}
			\centering
			\includegraphics[width=\textwidth]{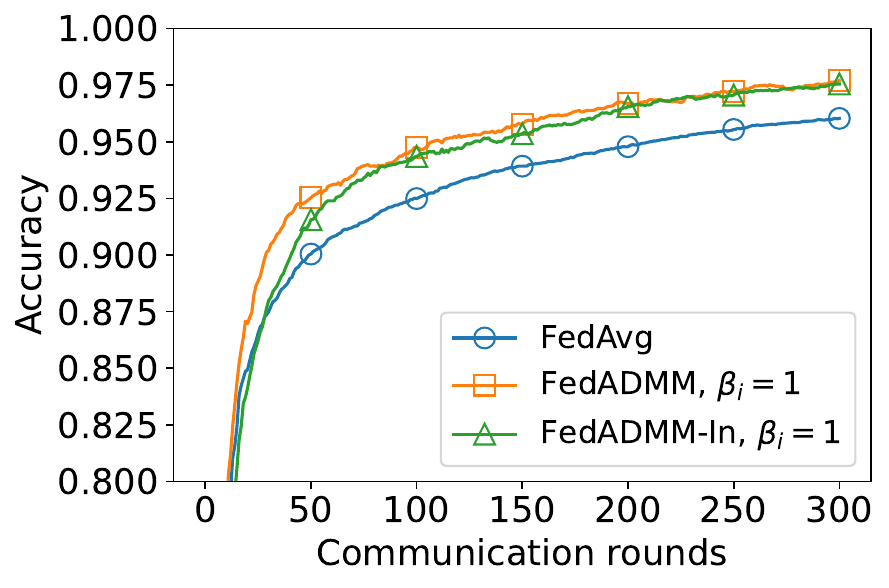}
			\caption{MNIST with CNN.}
			\label{fig_cmp_in_b}
		\end{subfigure}
		\hspace{10pt}
		\begin{subfigure}[b]{0.3\textwidth}
			\centering
			\includegraphics[width=\textwidth]{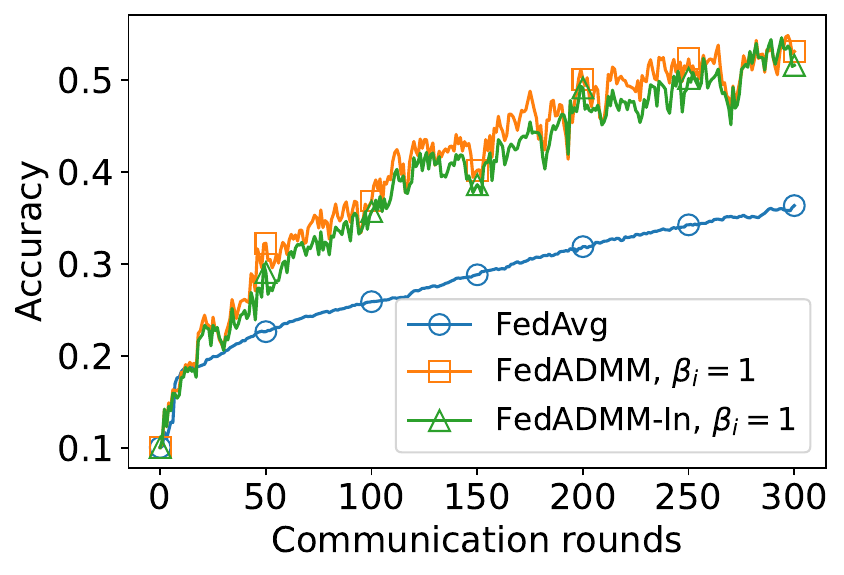}
			\caption{CIFAR-10 with ResNet.}
			\label{fig_cmp_in_c}
		\end{subfigure}
		
		\begin{subfigure}[b]{0.3\textwidth}
			\centering
			\includegraphics[width=\textwidth]{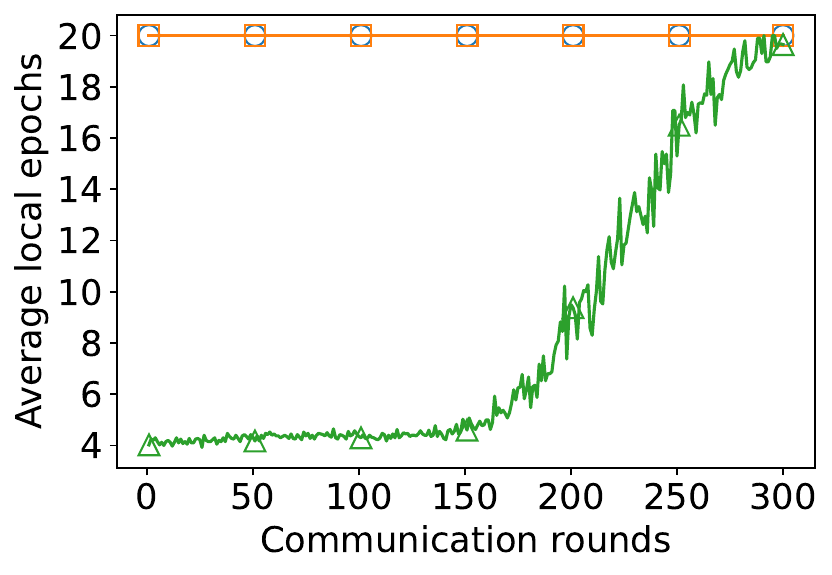}
			\caption{Linear regression.}
			\label{fig_cmp_in_d}
		\end{subfigure}
		\hspace{10pt}
		\begin{subfigure}[b]{0.31\textwidth}
			\centering
			\includegraphics[width=\textwidth]{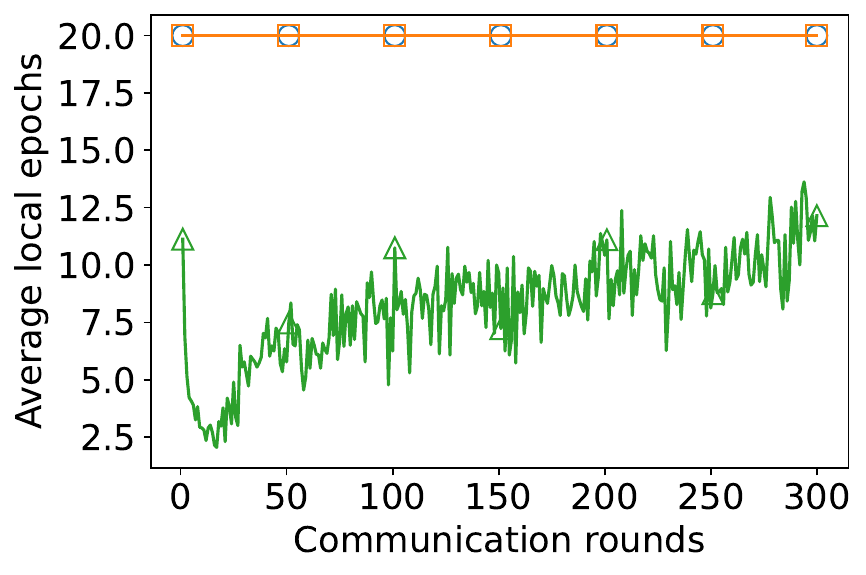}
			\caption{MNIST with CNN.}
			\label{fig_cmp_in_e}
		\end{subfigure}
		\hspace{10pt}
		\begin{subfigure}[b]{0.3\textwidth}
			\centering
			\includegraphics[width=\textwidth]{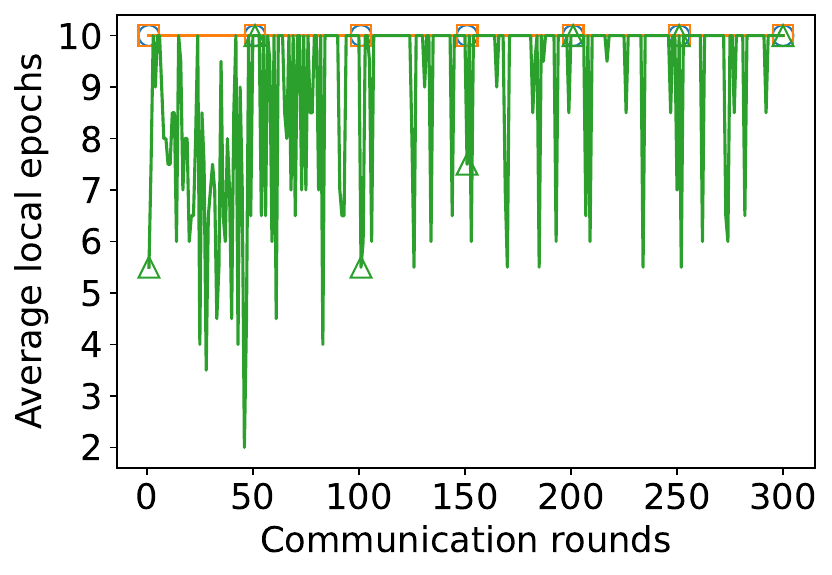}
			\caption{CIFAR-10 with ResNet.}
			\label{fig_cmp_in_f}
		\end{subfigure}
		\caption{{Comparison results of FedADMM-In in three examples.}
		}
		\label{fig_cmp_in}
	\end{figure}

	In this subsection, we compare our FedADMM-In algorithm with the vanilla FedADMM and FedAvg algorithms. The results with the penalty parameter $\beta_i=1$ are plotted in \Cref{fig_cmp_in}, while other results are summarized in \Cref{tab_results_three_egs}.
	
	As shown in Figures \ref{fig_cmp_in_a}-\ref{fig_cmp_in_c}, FedADMM-In achieves superior results compared to FedAvg and comparable results to FedADMM across all three examples. Furthermore, Figures \ref{fig_cmp_in_d}-\ref{fig_cmp_in_f} demonstrate that FedADMM-In allows clients to execute fewer local epochs compared to the other two algorithms while maintaining training performance. As detailed in \Cref{tab_results_three_egs}, FedADMM-In achieves an average local epoch reduction of 35.4\%, 49.0\%, and 16.6\% in Examples 1-3, respectively, leading to a substantial improvement in computational savings. These results indicate that our proposed FedADMM-In algorithm can achieve comparable or superior training outcomes while reducing the computational load on clients, significantly mitigating the potential waste of valuable computational resources in FL.
	
	\subsubsection{Evaluation of FedADMM-InSa}
	
\begin{figure}[!ht]
		\centering
		\begin{subfigure}[b]{0.33\textwidth}
			\centering
			\includegraphics[width=\textwidth]{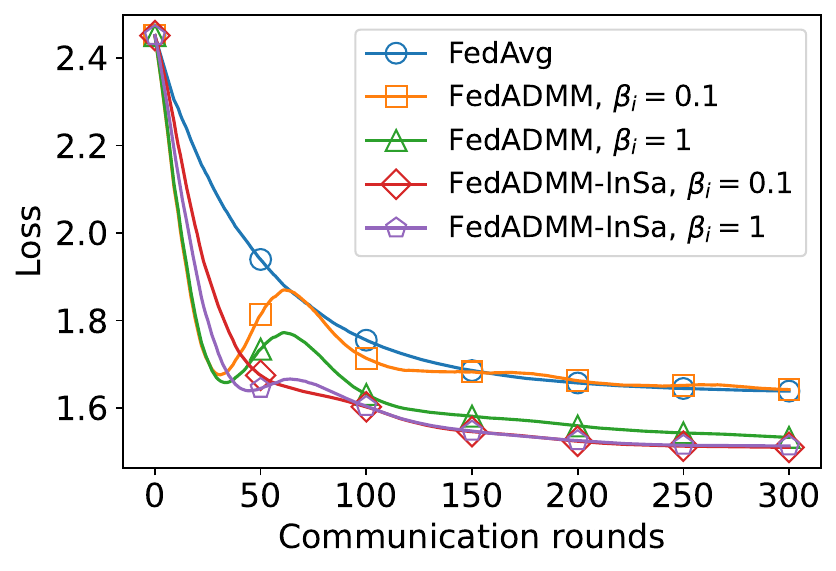}
			\caption{Training loss.}
			\label{fig_cmp_eg1_a}
		\end{subfigure}%
		\begin{subfigure}[b]{0.33\textwidth}
			\centering
			\includegraphics[width=\textwidth]{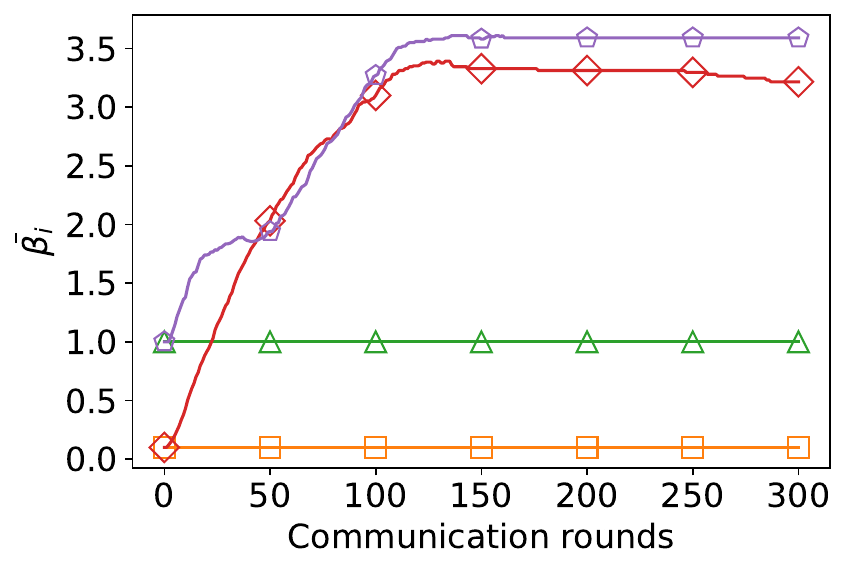}
			\caption{Average penalty parameters.}
			\label{fig_cmp_eg1_b}
		\end{subfigure}
		\begin{subfigure}[b]{0.33\textwidth}
			\centering
			\includegraphics[width=\textwidth]{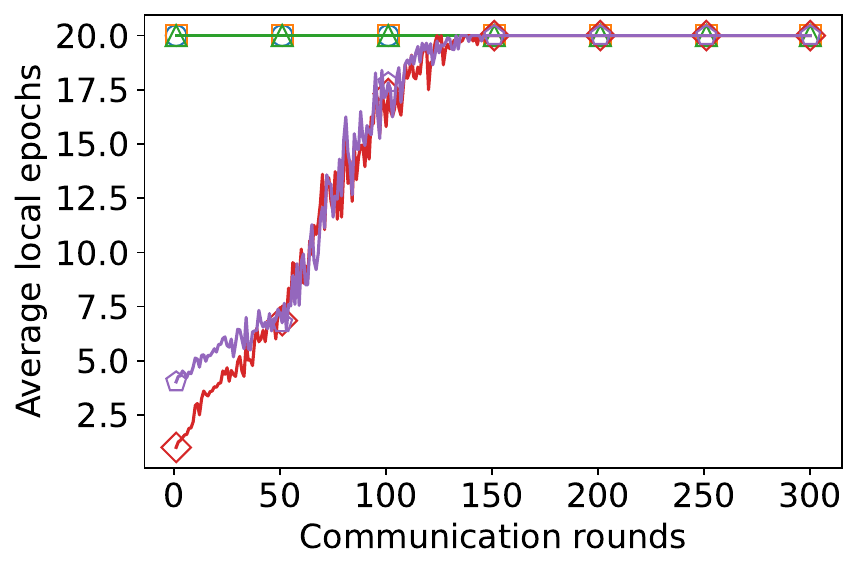}
			\caption{Average local epochs.}
			\label{fig_cmp_eg1_c}
		\end{subfigure}
		\caption{{Comparison results of FedADMM-InSa in Example 1 (Linear regression). }
		}
		\label{fig_cmp_eg1}
	\end{figure}

\begin{figure}[!ht]
		\centering
		\begin{subfigure}[b]{0.41\textwidth}
			\centering
			\includegraphics[width=\textwidth]{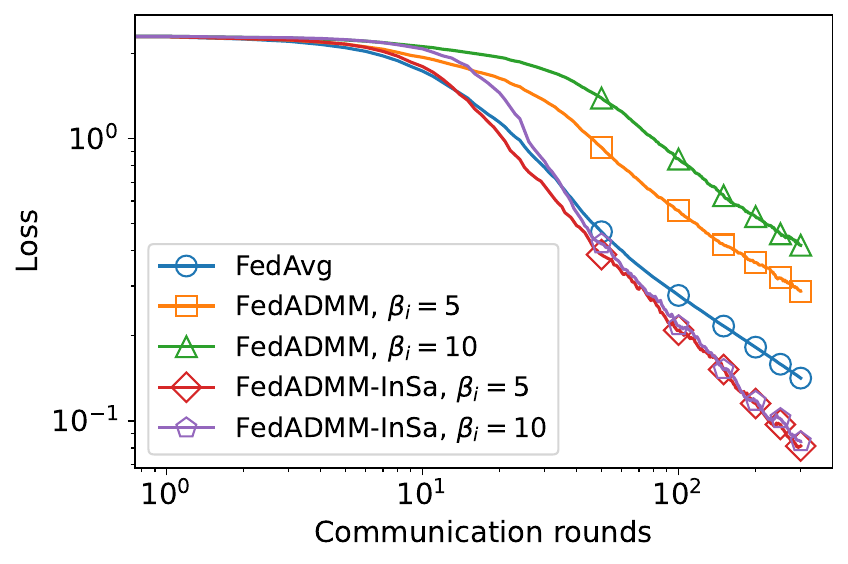}
			\caption{Training loss.}
			\label{fig_cmp_eg2_a}
		\end{subfigure}
		\begin{subfigure}[b]{0.41\textwidth}
			\centering
			\includegraphics[width=\textwidth]{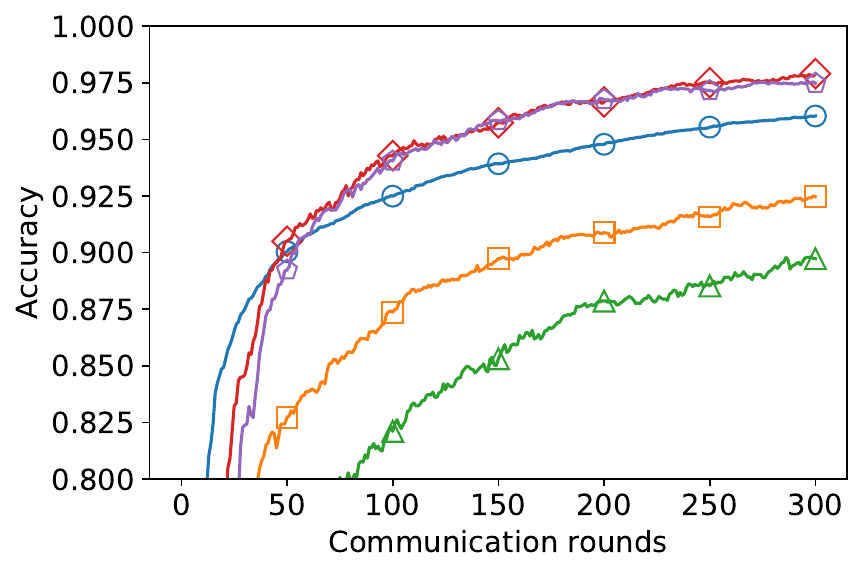}
			\caption{Test accuracy.}
			\label{fig_cmp_eg2_b}
		\end{subfigure}
		
		\begin{subfigure}[b]{0.41\textwidth}
			\centering
			\includegraphics[width=\textwidth]{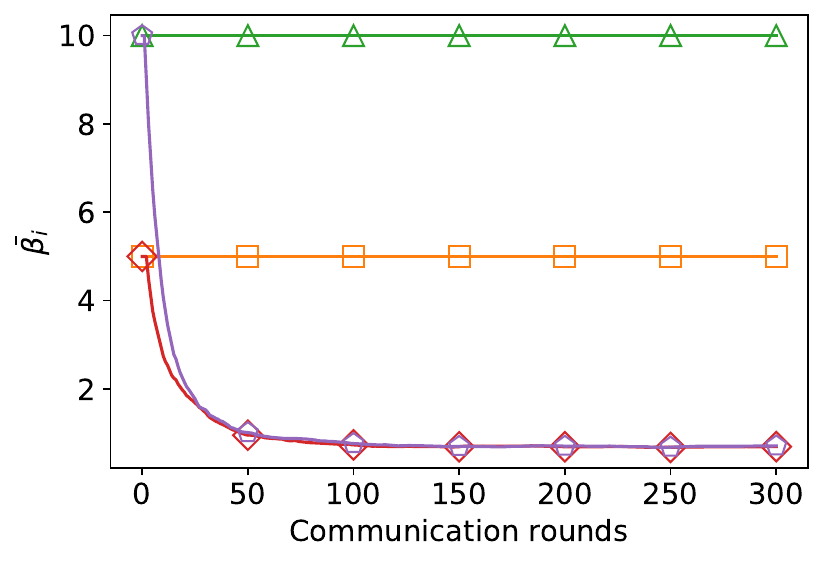}
			\caption{Average penalty parameters.}
			\label{fig_cmp_eg2_c}
		\end{subfigure}
		\begin{subfigure}[b]{0.41\textwidth}
			\centering
			\includegraphics[width=\textwidth]{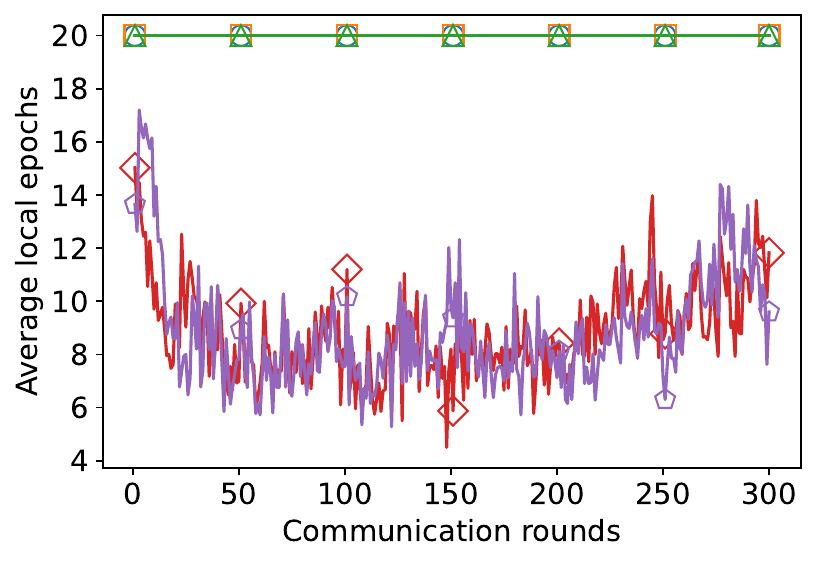}
			\caption{Average local epochs.}
			\label{fig_cmp_eg2_d}
		\end{subfigure}
		\caption{{Comparison results of FedADMM-InSa in Example 2 (MNIST with CNN).} 
		}
		\label{fig_cmp_eg2}
	\end{figure}

\begin{figure}[!t]
		\centering
		\begin{subfigure}[b]{0.415\textwidth}
			\centering
			\includegraphics[width=\textwidth]{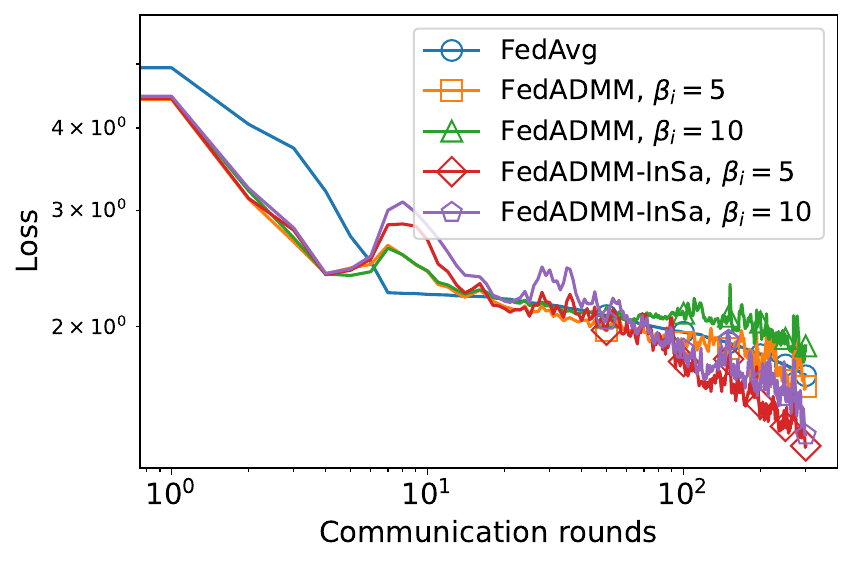}
			\caption{Training loss.}
			\label{fig_cmp_eg3_a}
		\end{subfigure}
		\begin{subfigure}[b]{0.405\textwidth}
			\centering
			\includegraphics[width=\textwidth]{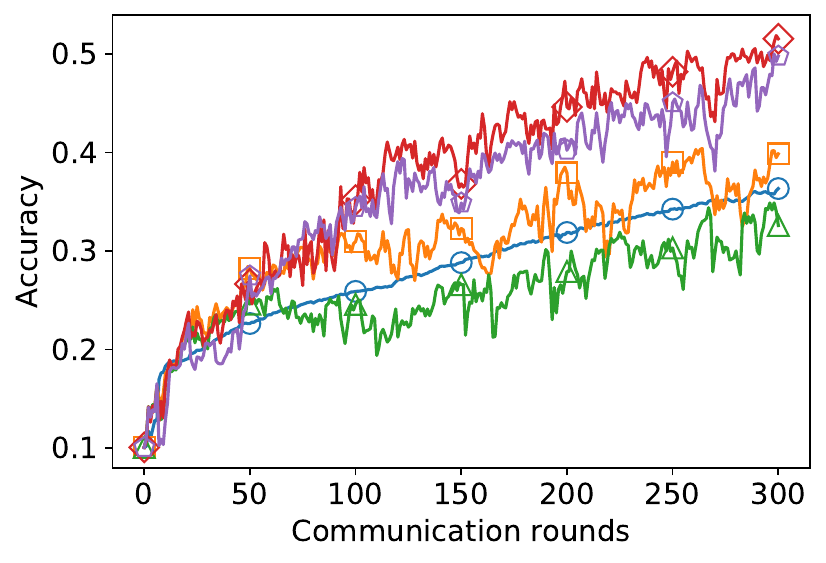}
			\caption{Test accuracy.}
			\label{fig_cmp_eg3_b}
		\end{subfigure}
		
		\begin{subfigure}[b]{0.41\textwidth}
			\centering
			\includegraphics[width=\textwidth]{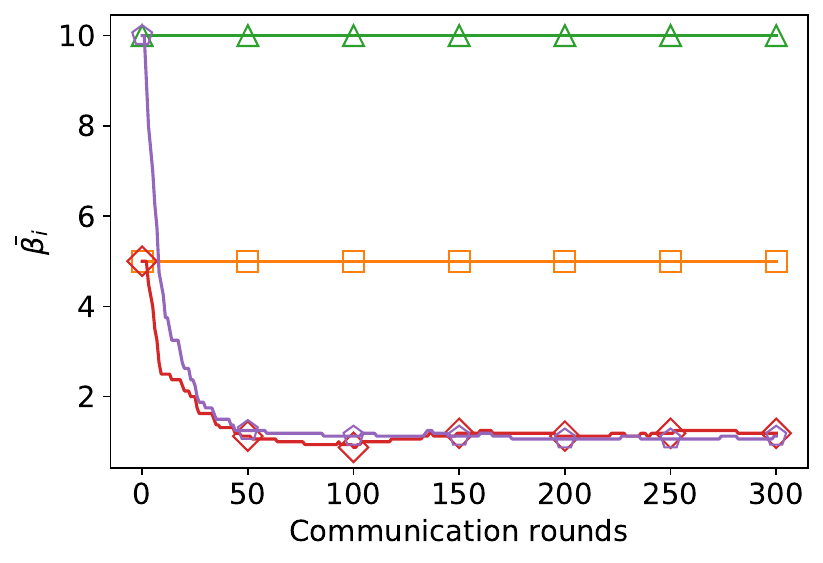}
			\caption{Average penalty parameters.}
			\label{fig_cmp_eg3_c}
		\end{subfigure}
		\begin{subfigure}[b]{0.41\textwidth}
			\centering
			\includegraphics[width=\textwidth]{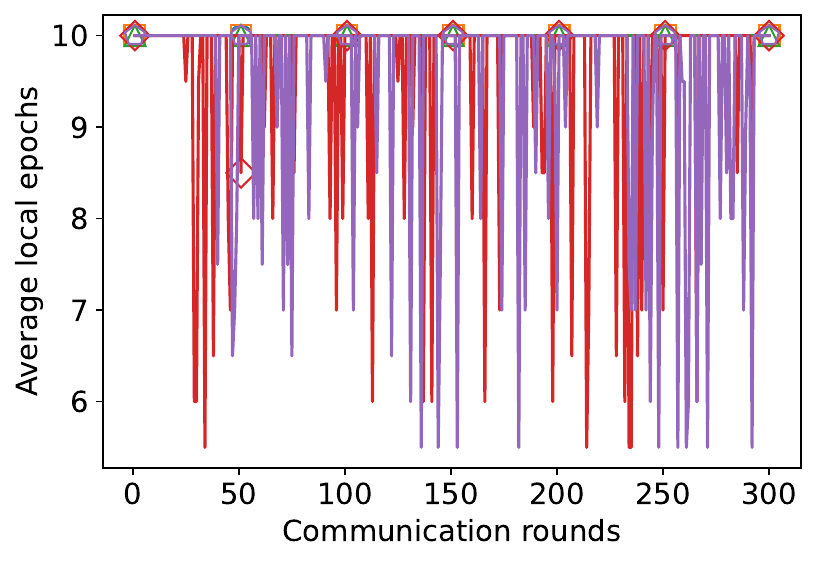}
			\caption{Average local epochs.}
			\label{fig_cmp_eg3_d}
		\end{subfigure}
		\caption{{Comparison results of FedADMM-InSa in Example 3 (CIFAR-10 with ResNet).} 
		}
		\label{fig_cmp_eg3}
	\end{figure}

	In this subsection, we present comparison results of our FedADMM-InSa algorithm with the vanilla FedADMM and FedAvg algorithms. The results for the three examples are presented in Figures \ref{fig_cmp_eg1}-\ref{fig_cmp_eg3} and are also summarized in \Cref{tab_results_three_egs}.
	
	\paragraph{Results of Example 1 (Linear regression)}
	\Cref{fig_cmp_eg1} shows the comparison results of our FedADMM-InSa with the vanilla FedADMM and FedAvg algorithms. We plot scenarios with $\beta_i \in \{0.1, 1\}$, and other results can be seen in \Cref{tab_results_three_egs}. It can be seen from \Cref{fig_cmp_eg1_a} that our FedADMM-InSa consistently achieves the lowest training loss. For FedADMM, it performs better than FedAvg when $\beta_i=1$. However, for $\beta_i=0.1$, FedADMM only decreases to a loss value similar to that of FedAvg, while our FedADMM-InSa reaches the lowest loss value. The average $\beta_i$ of clients are plotted in \Cref{fig_cmp_eg1_b}. For FedADMM-InSa, despite different initial $\beta_i$ values, they converge to a similar value around three. Additionally, \Cref{fig_cmp_eg1_c} presents the number of average local epochs of active clients in each round. It is evident that FedADMM-InSa requires fewer local epochs than FedADMM and FedAvg. In the later training rounds, the local epochs saturate probably because the algorithm has approached the local minimum.

	\paragraph{Results of Example 2 (MNIST with CNN)} \Cref{fig_cmp_eg2} shows the comparison results on the MNIST dataset using a CNN model. Similar to Example 1, the performance of FedADMM varies significantly with different $\beta_i$ values. In the two cases plotted with $\beta_i \in \{5, 10\}$, FedADMM performs even worse compared to FedAvg, highlighting its sensitivity to the choice of $\beta_i$. Conversely, our proposed FedADMM-InSa consistently achieves superior performance, demonstrating lower training loss and higher test accuracy across all algorithms.
	
	\Cref{fig_cmp_eg2_c} illustrates the change in average penalty parameters of all clients. It can be seen that FedADMM-InSa maintains an effective penalty parameter adaptation process, contributing to its robust performance. Meanwhile, \Cref{fig_cmp_eg2_d} shows the average local epochs. FedADMM-InSa requires much fewer local epochs on average compared to FedADMM, indicating a more efficient training process. This efficiency is critical in FL scenarios, as the client can complete local training faster without compromising the global model accuracy.
	
	\paragraph{Results of Example 3 (CIFAR-10 with ResNet-20)} 
	In \Cref{fig_cmp_eg3}, the performance of different algorithms on the CIFAR-10 dataset using a ResNet-20 model is presented. Similar to the previous two examples, FedADMM-InSa achieves the best performance, as evidenced by the lowest training loss in \Cref{fig_cmp_eg3_a} and the highest test accuracy in \Cref{fig_cmp_eg3_b}. The average local epochs required by FedADMM-InSa, as shown in \Cref{fig_cmp_eg3_d}, are still slightly lower than those conducted by FedADMM and FedAvg. The penalty parameter updates in \Cref{fig_cmp_eg3_c} illustrate the adaptive capability of FedADMM-InSa, contributing to its overall effectiveness. In summary, the comparison results across all three examples consistently demonstrate that our FedADMM-InSa outperforms both FedAvg and FedADMM in terms of test accuracy, workload reduction, and robustness across various penalty parameters.
	
	\subsubsection{Evaluation of the Self-Adaptive Penalty Parameter Scheme \eqref{self_adp_beta_rule}} 
	As observed in the three examples above, the pre-selected penalty parameter $\beta_i$ plays a crucial role in the performance of the FedADMM algorithm. For example, it performs better with a larger $\beta_i$ in Example 1, but with a smaller $\beta_i$ in Examples 2 and 3. In this subsection, we compare the results of FedADMM and FedADMM-InSa under different values of $\beta_i \in \{0.1, 1, 2, 5, 10\}$ using Example 2. The results of the other two examples are similar and omitted to save space.
	
\begin{figure}[!t]
		\centering
		\begin{subfigure}[b]{0.33\textwidth}
			\centering
			\includegraphics[width=\textwidth]{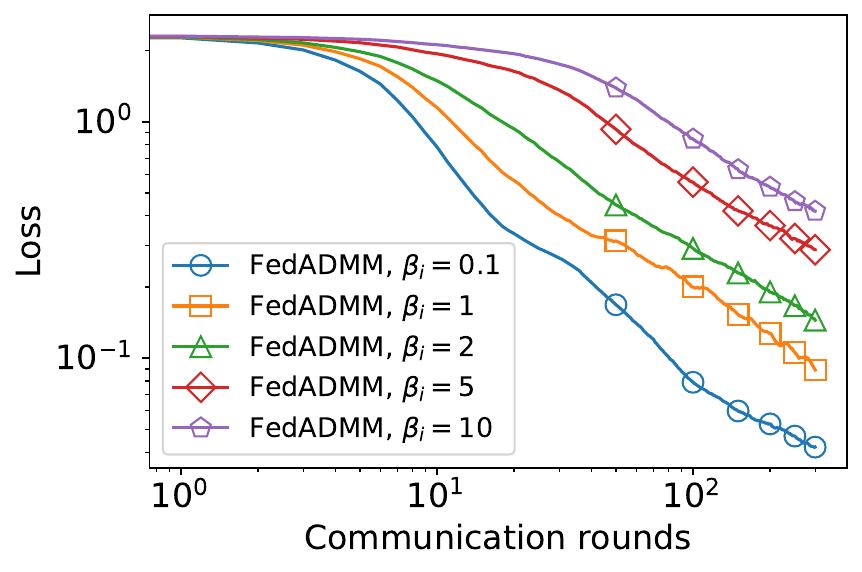}
			\caption{Training loss.}
			\label{fig_cmp_SA_a}
		\end{subfigure}	
		\begin{subfigure}[b]{0.33\textwidth}
			\centering
			\includegraphics[width=\textwidth]{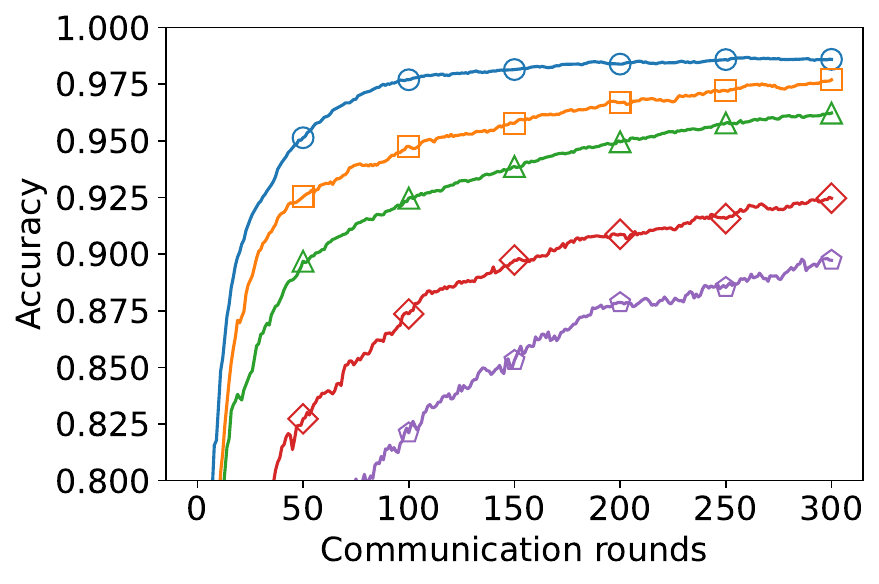}
			\caption{Test accuracy.}
			\label{fig_cmp_SA3_b}
		\end{subfigure}		
		\begin{subfigure}[b]{0.32\textwidth}
			\centering
			\includegraphics[width=\textwidth]{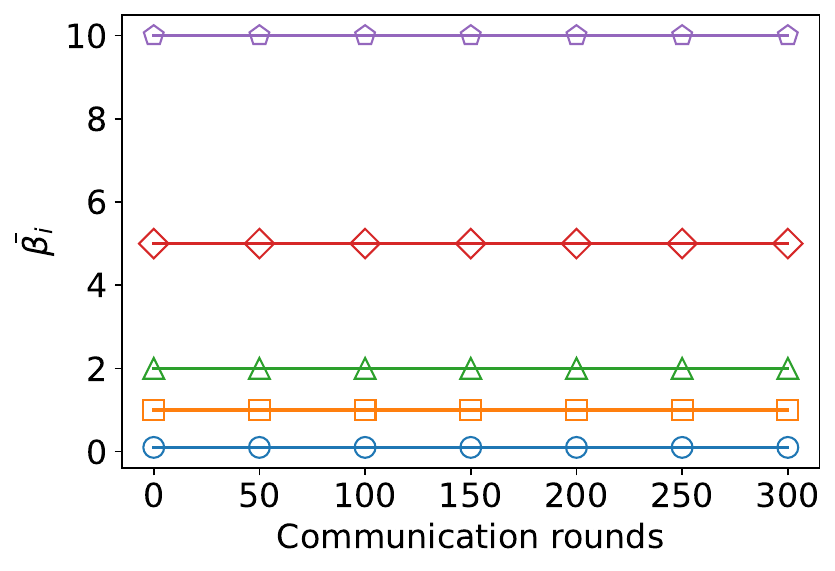}
			\caption{Average penalty parameters.}
			\label{fig_cmp_SA_c}
		\end{subfigure}
		
		\vspace{15pt}
		
		\begin{subfigure}[b]{0.33\textwidth}
			\centering
			\includegraphics[width=\textwidth]{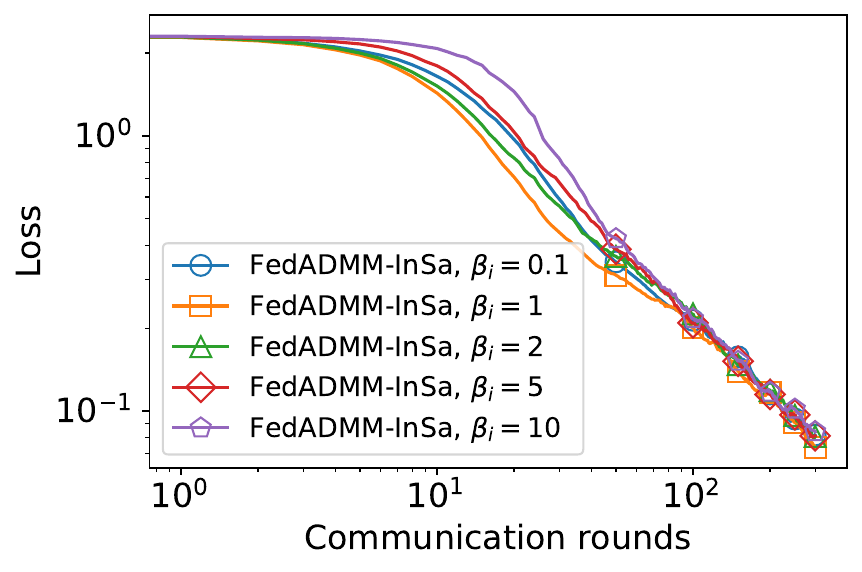}
			\caption{Training loss.}
			\label{fig_cmp_SA_d}
		\end{subfigure}
		\begin{subfigure}[b]{0.33\textwidth}
			\centering
			\includegraphics[width=\textwidth]{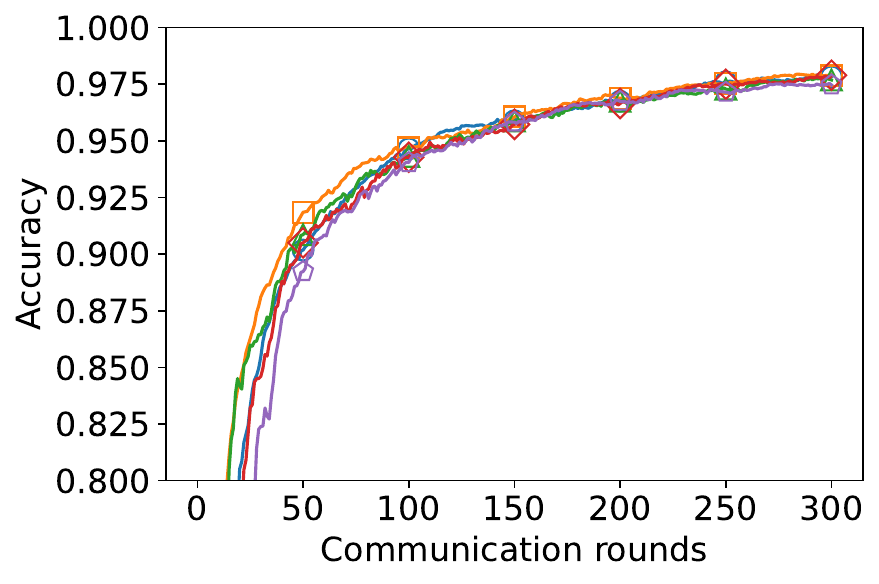}
			\caption{Test accuracy.}
			\label{fig_cmp_SA_e}
		\end{subfigure}
		\begin{subfigure}[b]{0.32\textwidth}
			\centering
			\includegraphics[width=\textwidth]{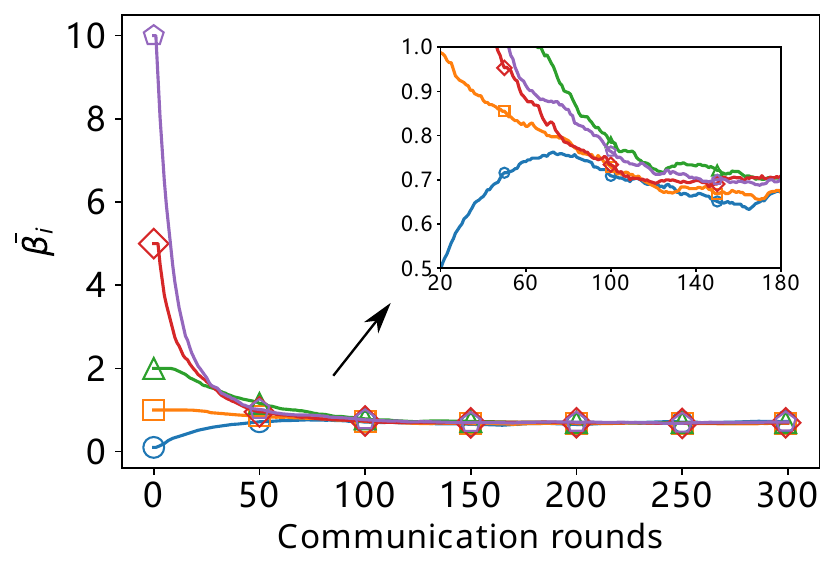}
			\caption{Average penalty parameters.}
			\label{fig_cmp_SA_f}
		\end{subfigure}
		\caption{{Comparison of FedADMM and FedADMM-InSa with different values of $\beta_i$ in Example 2. FedADMM (top row) uses fixed penalty parameters. Our FedADMM-InSa (bottom row) utilizes adaptive penalty parameter scheme \eqref{self_adp_beta_rule}.}}
		\label{fig_cmp_SA}
	\end{figure}

	It is clear from \Cref{fig_cmp_SA} that our proposed FedADMM-InSa (bottom row) demonstrates robustness across different values of $\beta_i$, while FedADMM (top row) only performs well under certain values. As shown in \Cref{fig_cmp_SA_f}, our proposed adaptive scheme dynamically adjusts the penalty parameter and ensures it converges to stable values, allowing our algorithm to maintain robust performance across different initial penalty parameter values. This adaptive mechanism prevents over-penalization, which can hinder convergence, and under-penalization, which can lead to insufficient coordination among clients. As a result, as shown in \Cref{fig_cmp_SA_d} and \Cref{fig_cmp_SA_e}, FedADMM-InSa achieves lower training loss and higher test accuracy. Moreover, FedADMM-InSa also requires fewer local epochs compared to FedADMM, as detailed in \Cref{tab_results_three_egs}.
	Overall, our adaptive penalty parameter scheme \eqref{self_adp_beta_rule} enhances the flexibility and robustness of our FedADMM-InSa algorithm, making it a superior choice for FL applications where training efficiency and computational resources are critical considerations.
	
\section{Conclusions}\label{sec_conclusions}
In this paper, we introduce the FedADMM-InSa algorithm, a novel approach that leverages the alternating direction method of multipliers (ADMM) to address the challenges of federated learning (FL) in the presence of data and system heterogeneity. Distinguished from current FedADMM methods, our inexact and self-adaptive algorithm mitigates the need for intricate empirical hyperparameter settings. The introduced inexactness criterion improves computational efficiency by removing the requirement to determine local training accuracy in advance. Additionally, the self-adaptive scheme dynamically adjusts each client's penalty parameter, enhancing the robustness of our algorithm. Numerical tests demonstrate the reduction in clients' local computational load and accelerated learning performance on both synthetic and real-world datasets, highlighting the practical advancements our approach brings to FL systems. In addition to the demonstrated benefits of our proposed algorithm, the integration of privacy-preserving techniques and the investigation of the algorithm's performance in large-scale FL applications are interesting and challenging future directions.

\section*{Acknowledgments}
Authors' names are listed in alphabetical order by family name to signify equal contributions. The authors are grateful to anonymous referees for their valuable comments which have helped improve the paper substantially.
This work has been funded by the Humboldt Research
Fellowship for postdoctoral researchers, the Alexander
von Humboldt-Professorship program, the European Union's Horizon Europe MSCA project Modelling and control of flexible structures interacting with fluids (ModConFlex; grant agreement No. 101073558; call HORIZON-MSCA-2021-DN-01-01; project DOI: \href{https://doi.org/10.3030/101073558}{10.3030/101073558}), the COST Action MAT-DYN-NET, the Transregio 154 Project of the DFG, grants PID2020-112617GB-C22 and TED2021-131390B-I00 of MINECO (Spain).
Madrid Government - UAM Agreement for the Excellence of the University Research Staff in the context of the V PRICIT (Regional Programme of Research and Technological Innovation).

\renewcommand{\thesection}{\Alph{section}}
\setcounter{section}{0}
\section{Appendix}\label{sec_appendix}

\subsection{Convergence Analysis of the Sequence Generated by \eqref{fedadmm_in}}\label{sec_convergence}
\subsubsection{Notations}\label{subsec_preparations}
To present our convergence analysis in a compact form, we first rewrite the augmented Lagrangian function \eqref{AL_sum}  as
\begin{equation}\label{aug_lag_vector}
	\begin{aligned}
		L_{\alpha, \beta}(u,\lambda,z) 
		&= \sum_{i=1}^m {\alpha_i} \left(f_i(u_i) - \lambda_i^{\top}\left(u_i-z\right) + \frac{\beta_i}{2}\left\|u_i-z\right\|^2\right) \\
		& = \alpha^{\top} {f}(u)-\lambda^{\top} I_\alpha(u-B z)+\frac{1}{2}\|u-B z\|_{I \alpha I_\beta}^2.
	\end{aligned}
\end{equation}
Here, $\lambda_i \in \mathbb{R}^n$ is the Lagrange multiplier associated with the equality constraint $u_i=z$, $\beta_i>0$ is a penalty parameter, 
\begin{equation}\label{notations}
	\begin{aligned}
		& u=\left(u_1, \ldots ,u_m\right)^{\top} \in \mathbb{R}^{m n}, \
		\lambda=\left(\lambda_1, \ldots, \lambda_m\right)^{\top} \in \mathbb{R}^{m n},\\
		& \alpha=\left(\alpha_1, \ldots ,\alpha_m\right)^{\top} \in \mathbb{R}^m, \
		\beta=\left(\beta_1, \ldots ,\beta_m\right)^{\top} \in \mathbb{R}^m, \\
		& f(u) \coloneqq  \left(f_1\left(u_1\right), \ldots, f_m\left(u_m\right)\right)^{\top} \in \mathbb{R}^m,\\
		& \nabla f(u) \coloneqq  \left(\nabla f_1\left(u_1\right), \ldots, \nabla f_m\left(u_m\right)\right)^{\top} \in \mathbb{R}^{m n},\\
		& e^k(u) \coloneqq  (e_1^k(u_1),\ldots, e_m^k(u_m))^{\top} \in \mathbb{R}^{m n},\\
		& B=(\underbrace{I_n, \ldots ,I_n}_m)^{\top} \in \mathbb{R}^{m n \times n},\\
	\end{aligned}
\end{equation}
and diagonal matrices $I_\alpha, I_\beta \in \mathbb{R}^{m n \times m n}$ are
\begin{equation}
	I_\alpha =\left(\begin{array}{lll}
		\alpha_1 I_n & & \\
		& \ddots & \\
		& & \alpha_m I_n
	\end{array}\right), \
	I_\beta =\left(\begin{array}{lll}
		\beta_1 I_n & & \\
		& \ddots & \\
		& & \beta_m I_n
	\end{array}\right).
\end{equation}
We define the $H$-norm with a symmetric and positive-definite matrix $H \in \mathbb{R}^{n \times n}$ as 
\begin{equation}\label{H_norm}
	\|x\|_H= (x^{\top} H x)^{1/2}, \quad \forall x \in \mathbb{R}^{n}.
\end{equation}
We also denote $w \in W\coloneqq  \mathbb{R}^{m n} \times \mathbb{R}^{m n} \times \mathbb{R}^{n}, v \in V\coloneqq  \mathbb{R}^{m n} \times \mathbb{R}^{n}$ and the function $F(w)$ as follows:
\begin{equation}\label{F_of_w}\texttt{}
	w=\left(\begin{array}{c}
		u \\
		\lambda \\
		z
	\end{array}\right), \ v=\left(\begin{array}{c}
		\lambda \\
		z
	\end{array}\right), \ F(w)=\left(\begin{array}{c}
		I_\alpha (\nabla f(u)-\lambda) \\
		I_\alpha (u-B z) \\
		B^{\top} I_\alpha \lambda
	\end{array}\right).
\end{equation}
Recall that $(u^{*}, z^{*})$ is the unique solution of problem \eqref{prob_FL_consensus} and $\lambda^{*}$ is the solution of its dual problem. Let $w^{*} =(u^{*}, \lambda^{*}, z^{*})^{\top} $, then the first-order optimality conditions of problem \eqref{prob_FL_consensus} and its dual problem can be expressed as:
\begin{equation}\label{VI_opt}
	F(w^*) = 0.
\end{equation}
From Assumption \ref{ass_strongly_convex} and the formula of $F$, it is easy to deduce that the solution $w^*$ is unique.

\subsubsection{Optimality Conditions}
Using the notations in Appendix \ref{subsec_preparations}, we can rewrite scheme \eqref{fedadmm_in} as 
\begin{subequations}\label{fedadmm}
	\begin{empheq}[left=\empheqlbrace]{align}
		&u^{k+1} \approx \arg \min_{u \in \mathbb{R}^{mn}}  L_{\alpha, \beta}\left(u, \lambda^k, z^k\right), \label{fedadmm_u}
		\\
		&\lambda^{k+1}=\lambda^k-I_\beta\left(u^{k+1}-B z^k\right),\label{fedadmm_lambda}
		\\ 
		&z^{k+1}=\arg \min_{z \in \mathbb{R}^n} L_{\alpha, \beta}\left(u^{k+1}, \lambda^{k+1}, z\right)+\frac{\delta}{2}\left\|B(z-z^k)\right\|_{I_\alpha I_\beta}^2. \label{fedadmm_z}
	\end{empheq}
\end{subequations}
For the point $w^{k+1}=\left(u^{k+1}, z^{k+1}, \lambda^{k+1}\right)^{\top}$ generated by \eqref{fedadmm}, utilizing the definitions of $e_i^k(u_i)$ in \eqref{error_i_k} and $e^k(u)$ in \eqref{notations}, and combining the first-order optimality condition of \eqref{fedadmm_z}, we have:
\begin{subequations}
	\begin{empheq}[left=\empheqlbrace]{align}
		&\nabla_u L_{\alpha, \beta}\left(u^{k+1}, \lambda^k, z^k\right) 
		= I_\alpha\left(e^k(u^{k+1})\right) \label{opt_1}, \\
		&\lambda^{k+1}=\lambda^k-I_\beta\left(u^{k+1}-B z^k\right) \label{opt_2},\\
		&\left(z-z^{k+1}\right)^{\top}\left(B^{\top} I_\alpha \lambda^{k+1}-B^{\top} I_\alpha I_\beta(u^{k+1}-B z^{k+1})+\delta B^{\top} I_\alpha I_\beta B(z^{k+1}-z^k)\right) \geq 0. \label{opt_3}
	\end{empheq}
\end{subequations}
Moreover, note that the inexactness condition \eqref{sigma} implies that
\begin{equation}
	0<\frac{\sigma_i^2\beta_i}{2c_i(1-\sigma_i)^2}=\left(\frac{\sigma_i}{2c_i(1-\sigma_i)}\right)\left(\frac{\sigma_i \beta_i}{1-\sigma_i}\right)<1, \quad \forall i \in [m].
\end{equation}
Then, there exists a constant $\mu_i>0$ such that
\begin{equation}\label{mu_positive}
	\left(c_i - \frac{\mu_i}{2} \frac{\sigma_i}{1-\sigma_i}\right)>0 \quad \text { and } \quad\left(1-\frac{1}{\mu_i} \frac{\sigma_i\beta_i}{1-\sigma_i}\right)>0, \quad \forall i \in [m].
\end{equation}
The above inequalities will be used later in the proof.

\subsubsection{Convergence Proof}
With the above preparations, we start to prove the convergence of sequence $\left\{w^k\right\}$ generated by \eqref{fedadmm_in}. We first prove two lemmas which will be useful in the following discussion.
First of all, we analyze how different the point $w^k$ generated by our algorithm is away from the solution $w^*$ of \eqref{VI_opt}.
\begin{lemma}\label{lemma1}
	Let $\{w^k\}=\{(u^k, \lambda^k, z^k)^{\top}\}$ be the sequence generated by scheme \eqref{fedadmm_in}. Then, for all $w \in W$, one has
	\begin{equation}\label{res_lemma1}
		\begin{aligned}
			\left(w^{k+1}-w\right)^{\top} F\left(w^{k+1}\right) 
			\leq& \left(u^{k+1}-u\right)^{\top} \nabla_u L_{\alpha, \beta} \left(u^{k+1}, \lambda^k, z^k\right)\\
			&+\frac{1}{2}\left(
			\|v^k-v\|_{H_1}^2 
			-\|v^{k+1}-v\|_{H_1}^2
			-\|v^k-v^{k+1}\|_{H_1}^2\right),
		\end{aligned}
	\end{equation}
	where 
	\begin{equation}\label{H1}
		v=\left(\begin{array}{l}
			\lambda \\
			z
		\end{array}\right), \
		H_1=\left(\begin{array}{ll}
			I_\alpha I_\beta^{-1} & I_\alpha B \\
			B^{\top} I_\alpha & (1+\delta) B^{\top} I_\alpha I_\beta B
		\end{array}\right) \succ 0.
	\end{equation}
\end{lemma}
\begin{proof}
	For all $w \in W$, we have
	\begin{equation}\label{lemma1_VI_tmp1}
		\begin{aligned}
			\left(w^{k+1}-w\right)^{\top} F\left(w^{k+1}\right)
			\overset{\eqref{F_of_w}}{=}& \left(u^{k+1}-u\right)^{\top} I_\alpha\left(\nabla f(u^{k+1})-\lambda^{k+1}\right)\\
			&+\left(\lambda^{k+1}-\lambda\right)^{\top} I_\alpha\left(u^{k+1}-B z^{k+1}\right) \\
			&+\left(z^{k+1}-z\right)^{\top} B^{\top} I_\alpha \lambda^{k+1}
		\end{aligned}
	\end{equation}
	It follows from \eqref{aug_lag_vector} that
	\begin{equation}\label{nabla_u_L_beta}
		\begin{aligned}
			\nabla_u L_{\alpha, \beta}\left(u^{k+1}, \lambda^k, z^k\right) 
			&= I_\alpha\left(\nabla f(u^{k+1})-\lambda^k+I_\beta(u^{k+1}-B z^k)\right)\\
			&\overset{\eqref{opt_2}}{=} I_\alpha\left(\nabla f(u^{k+1})-\lambda^{k+1}\right).
		\end{aligned}
	\end{equation}
	For the first term on the right-hand side of \eqref{lemma1_VI_tmp1}, it follows from \eqref{nabla_u_L_beta} that
	\begin{equation}\label{lemma1_VI_1}
		\begin{aligned}
			& \left(u^{k+1}-u\right)^{\top} I_\alpha\left(\nabla f(u^{k+1})-\lambda^{k+1}\right)
			\overset{\eqref{nabla_u_L_beta}}{=} \left(u^{k+1}-u\right)^{\top} \nabla_u L_{\alpha, \beta}\left(u^{k+1}, \lambda^k, z^k\right)\\
		\end{aligned}
	\end{equation}
	For the second term on the right-hand side of \eqref{lemma1_VI_tmp1}, we have
	\begin{equation}\label{lemma1_VI_2}
		\begin{aligned}
			&\left(\lambda^{k+1}-\lambda\right)^{\top} I_\alpha\left(u^{k+1}-B z^{k+1}\right)\\
			=&\left(\lambda^{k+1}-\lambda\right)^{\top} I_\alpha\left(u^{k+1} - Bz^{k} + Bz^{k} - Bz^{k+1}\right)\\
			=&\left(\lambda^{k+1}-\lambda\right)^{\top} I_\alpha\left(u^{k+1}-B z^k\right) + \left(\lambda^{k+1}-\lambda\right)^{\top} I_\alpha B\left(z^k-z^{k+1}\right)\\
			\overset{\eqref{opt_2}}{=} & \left(\lambda^{k+1}-\lambda\right)^{\top} I_\alpha I_{\beta}^{-1}\left(\lambda^k-\lambda^{k+1}\right) + \left(\lambda^{k+1}-\lambda\right)^{\top} I_\alpha B\left(z^k-z^{k+1}\right)
		\end{aligned}
	\end{equation}
	For the third term on the right-hand side of \eqref{lemma1_VI_tmp1}, we have
	\begin{equation}\label{lemma1_VI_3}
		\begin{aligned}
			& \left(z^{k+1}-z\right)^{\top} B^{\top} I_\alpha \lambda^{k+1} \\
			\overset{\eqref{opt_3}}{\leq}& 
			\left(z^{k+1}-z\right)^{\top} B^{\top} I_\alpha I_\beta\left(u^{k+1}-B z^{k+1}\right) + \delta \left(z^{k+1}-z\right)^{\top} B^{\top} I_\alpha I_\beta B\left(z^k-z^{k+1}\right) \\
			= & \left(z^{k+1}-z\right)^{\top} B^{\top} I_\alpha I_\beta\left(u^{k+1} - B z^{k} + B z^{k}-B z^{k+1}\right)\\
			&+\left(z^{k+1}-z\right)^{\top} \delta B^{\top} I_\alpha I_\beta B\left(z^k-z^{k+1}\right) \\
			= & \left(z^{k+1}-z\right)^{\top} B^{\top} I_\alpha I_\beta\left(u^{k+1} - B z^{k}\right) + \left(z^{k+1}-z\right)^{\top} B^{\top} I_\alpha I_\beta B \left( z^{k}-z^{k+1}\right)\\
			& + \delta \left(z^{k+1}-z\right)^{\top} B^{\top} I_\alpha I_\beta B\left(z^k-z^{k+1}\right) \\
			\overset{\eqref{opt_2}}{=} &
			\left(z^{k+1}-z\right)^{\top} B^{\top} I_\alpha\left(\lambda^k-\lambda^{k+1}\right) 
			+ \left(1+\delta\right)\left(z^{k+1}-z\right)^{\top} B^{\top} I_\alpha I_\beta B\left(z^k-z^{k+1}\right).
		\end{aligned}
	\end{equation}
	Substituting \eqref{lemma1_VI_1}-\eqref{lemma1_VI_3} back into \eqref{lemma1_VI_tmp1} yields
	\begin{equation}\label{lemma1_VI_tmp2}
		\begin{aligned}
			& \left(w^{k+1}-w\right)^{\top} F\left(w^{k+1}\right) \\
			\leq & \left(u^{k+1}-u\right)^{\top} \nabla_u L_{\alpha, \beta}\left(u^{k+1}, \lambda^k, z^k\right)\\
			&+\left(\lambda^{k+1}-\lambda\right)^{\top} I_\alpha I_{\beta}^{-1}\left(\lambda^k-\lambda^{k+1}\right)
			+\left(\lambda^{k+1}-\lambda\right)^{\top} I_\alpha B\left(z^k-z^{k+1}\right)\\
			&+\left(z^{k+1}-z\right)^{\top} B^{\top} I_\alpha\left(\lambda^k-\lambda^{k+1}\right)
			+ \left(1+\delta\right)\left(z^{k+1}-z\right)^{\top} B^{\top} I_\alpha I_\beta B\left(z^k-z^{k+1}\right).
		\end{aligned}
	\end{equation}
	Using $v$ and $H_1$ defined by \eqref{H1}, we can rewrite \eqref{lemma1_VI_tmp2} in a more compact form:
	\begin{equation}\label{lemma1_VI_tmp3}
		\begin{aligned}
			\left(w^{k+1}-w\right)^{\top} F\left(w^{k+1}\right) 
			\leq & \left(u^{k+1}-u\right)^{\top} \nabla_u L_{\alpha, \beta}\left(u^{k+1}, \lambda^k, z^k\right) \\
			& + \left(v^{k+1}-v\right)^{\top} H_1 \left(v^{k}-v^{k+1}\right).
		\end{aligned}
	\end{equation}
	Applying the identity 
	\begin{equation}\label{acMbc}
		(c-a)^{\top} M(b-c)=\frac{1}{2}\left(\|b-a\|_M^2-\|c-a\|_M^2-\|b-c\|_M^2\right)
	\end{equation}
	to the second term on the right-hand side of \eqref{lemma1_VI_tmp3}, we have
	\begin{equation}\label{identity1}
		\begin{aligned}
			\left(v^{k+1}-v\right)^{\top} H_1 \left(v^{k}-v^{k+1}\right) 
			= \frac{1}{2}\left(
			\|v^k-v\|_{H_1}^2 
			-\|v^{k+1}-v\|_{H_1}^2
			-\|v^k-v^{k+1}\|_{H_1}^2\right).
		\end{aligned}
	\end{equation}
	Substituting \eqref{identity1} back into \eqref{lemma1_VI_tmp3} yields
	\begin{equation}
		\begin{aligned}
			\left(w^{k+1}-w\right)^{\top} F\left(w^{k+1}\right) 
			\leq & \left(u^{k+1}-u\right)^{\top} \nabla_u L_{\alpha, \beta}\left(u^{k+1}, \lambda^k,z^k\right)\\
			&+ \frac{1}{2}
			\left(
			\|v^k-v\|_{H_1}^2 
			-\|v^{k+1}-v\|_{H_1}^2
			-\|v^k-v^{k+1}\|_{H_1}^2
			\right).
		\end{aligned}
	\end{equation}
	We thus complete the proof.
\end{proof}
On the right-hand side of \eqref{res_lemma1}, the three quadratic terms are easily amenable to manipulation across various indicators through algebraic operations. However, it is less apparent how the cross-term can be controlled to demonstrate the convergence of the sequence $\{w^k\}$. Therefore, we study this term and demonstrate that the sum of these cross-terms over $K \geq 1$ iterations can be bounded by certain quadratic terms. This result is shown in the following \Cref{lemma2}.
\begin{lemma}\label{lemma2}
	Let $\{w^k\}=\{(u^k, \lambda^k, z^k)^{\top}\}$ be the sequence generated by scheme \eqref{fedadmm_in}. For any integer $K \geq 1$ and $\mu_i$ satisfying \eqref{mu_positive}, one has
	\begin{equation}\label{res_lemma2}
		\begin{aligned}
			&\sum_{k=1}^K\left(u^{k+1} - u\right)^{\top} \nabla_u L_{\alpha, \beta} \left(u^{k+1}, \lambda^k, z^k\right)\\
			\leq& \sum_{k=1}^K \sum_{i=1}^m \alpha_i\frac{\mu_i}{2} \frac{\sigma_i}{1-\sigma_i}\left\|u_i^{k+1}-u_i\right\|^2 
			+ \sum_{k=1}^{K-1} \sum_{i=1}^m \alpha_i\frac{1}{2 \mu_i} \frac{\sigma_i}{1-\sigma_i} \beta_i\left\|v_i^k-v_i^{k+1}\right\|_{H_{\beta_i}}^2 \\
			& +\sum_{i=1}^m \alpha_i \frac{1}{2 \mu_i} \frac{\sigma_i}{1-\sigma_i}\left(\left\|e_i^0\left(u_i^1\right)\right\|+\sqrt{\beta_i}\left\|v_i^0-v_i^1\right\|_{H_{\beta_i}}\right)^2, \quad \forall u \in \mathbb{R}^{m n},
		\end{aligned}
	\end{equation}
	where 
	\begin{equation}\label{H_beta_i}
		v_i=\left(\begin{array}{l}
			\lambda_i \\
			z
		\end{array}\right), \
		H_{\beta_i}=\left(\begin{array}{cc}
			\frac{1}{\beta_i} I_n & I_n \\
			I_n &  \left(1+\delta\right)\beta_i I_n
		\end{array}\right) \succ 0.
	\end{equation}
\end{lemma}
\begin{proof}
	For the residual $e_i^k(u_i), i \in [m]$, it follows from \eqref{error_i_k} that
	\begin{subequations}
		\begin{empheq}[left=]{align}
			e_i^k\left(u_i^k\right) &= \nabla f_i\left(u_i^k\right)-\lambda_i^k+\beta_i\left(u_i^k-z^k\right), \label{e_i_k}
			\\
			e_i^{k-1}\left(u_i^k\right) &=\nabla f_i\left(u_i^k\right)-\lambda_i^{k-1}+\beta_i\left(u_i^k-z^{k-1}\right). \label{e_i_k-1}
		\end{empheq}
	\end{subequations}
	Combining \eqref{e_i_k} and \eqref{e_i_k-1} we have
	\begin{equation}\label{residual_e_i_k}
		e_i^k(u_i^{k}) 
		= e_i^{k-1}(u_i^k) + \beta_i (z^{k-1} - z^k)+\lambda_i^{k-1} -\lambda_i^k.
	\end{equation}
	For the residual $e_i^k(u_i^{k+1}), i \in [m]$, it follows from \eqref{inexact} that
	\begin{equation}\label{eik1}
		\begin{aligned}
			\left\|e_i^k(u_i^{k+1})\right\| 
			\overset{\eqref{inexact}}{\leq} & \sigma_i\left\|e_i^k(u_i^k)\right\|\\
			\overset{\eqref{residual_e_i_k}}{=} &
			\sigma_i\left\|e_i^{k-1}(u_i^k)
			+ \beta_i (z^{k-1} - z^k)+\lambda_i^{k-1} -\lambda_i^k\right\| \\
			\leq& \sigma_i\left\|e_i^{k-1}(u_i^k)\right\|+\sigma_i \left\|\beta_i (z^{k} - z^{k-1})+\lambda_i^{k} -\lambda_i^{k-1}\right\|.
		\end{aligned}
	\end{equation}
	For the second term on the right-hand side of \eqref{eik1}, with any $\delta \geq 0$, we have
	\begin{align}\label{eik1_sec_rhs}
		& \sigma_i \left\|\beta_i (z^{k} - z^{k-1})+\lambda_i^{k} -\lambda_i^{k-1}\right\| \notag\\
		=& \sigma_i \left(\beta_i^2 \left\|z^k-z^{k-1}\right\|^2
		+\left\|\lambda_i^k-\lambda_i^{k-1}\right\|^2
		+2 \beta_i \left(z^k-z^{k-1}\right)^{\top}\left(\lambda_i^k-\lambda_i^{k-1}\right)\right)^{\frac{1}{2}} \notag\\
		=& \sigma_i \sqrt{\beta_i} \left(\beta_i \left\|z^k-z^{k-1}\right\|^2+\frac{1}{\beta_i}\left\|\lambda_i^k-\lambda_i^{k-1}\right\|^2+2\left(z^k-z^{k-1}\right)^{\top}\left(\lambda_i^k-\lambda_i^{k-1}\right)\right)^{\frac{1}{2}} \\
		\leq& \sigma_i \sqrt{\beta_i} \left(\left(1+\delta\right)\beta_i \left\|z^k-z^{k-1}\right\|^2+\frac{1}{\beta_i}\left\|\lambda_i^k-\lambda_i^{k-1}\right\|^2+2\left(z^k-z^{k-1}\right)^{\top}\left(\lambda_i^k-\lambda_i^{k-1}\right)\right)^{\frac{1}{2}}. \notag
	\end{align}
	Using the $H$-norm notation in \eqref{H_norm} with $v_i$ and $H_{\beta_i}$ defined by \eqref{H_beta_i}, we have
	\begin{equation}\label{ineq_h_beta}
		\begin{aligned}
			\left\|v_i^k-v_i^{k-1}\right\|_{H_{\beta_i}}^2 
			=& \left(1+\delta\right) \beta_i \left\|z^k-z^{k-1}\right\|^2
			+\frac{1}{\beta_i}\left\|\lambda_i^k-\lambda_i^{k-1}\right\|^2\\
			&+2\left(z^k-z^{k-1}\right)^{\top}\left(\lambda_i^k-\lambda_i^{k-1}\right).
		\end{aligned}
	\end{equation}
	Substituting \eqref{eik1_sec_rhs} back into \eqref{eik1} and using \eqref{ineq_h_beta}, we have
	\begin{equation}\label{eik2}
		\begin{aligned}
			\left\|e_i^k(u_i^{k+1})\right\| 
			\leq& \sigma_i\left\|e_i^{k-1}(u_i^k)\right\|+\sigma_i \sqrt{\beta_i}\left\|v_i^k-v_i^{k-1}\right\|_{H_{\beta_i}}\\
			\leq& \sum_{j=0}^{k-1} \sigma_i^{k-j} \sqrt{\beta_i}\left\|v_i^j-v_i^{j+1}\right\|_{H_{\beta_i}}+\sigma_i^k\left\|e_i^0(u_i^1)\right\| .
		\end{aligned}
	\end{equation}
	By using \eqref{opt_1} and \eqref{eik2}, for any $\mu_i>0$ satisfying \eqref{mu_positive} and $u \in \mathbb{R}^{m n}$, we have
	\begin{align}
		& \sum_{k=1}^K\left(u^{k+1}-u\right)^{\top}\nabla_u L_{\alpha, \beta}\left(u^{k+1}, z^k, \lambda^k\right) 
		\overset{\eqref{opt_1}}{=} 
		\sum_{k=1}^K\left(u^{k+1}-u\right)^{\top} I_\alpha e^k(u^{k+1}) \notag\\
		=&
		\sum_{k=1}^K \sum_{i=1}^m \alpha_i \left(u_i^{k+1}-u_i\right)^{\top}  e_i^k (u_i^{k+1}) 
		\leq 
		\sum_{k=1}^K \sum_{i=1}^m \alpha_i\left\|u_i^{k+1}-u_i\right\|\left\|e_i^k(u_i^{k+1})\right\| \notag\\
		\overset{\eqref{eik2}}{\leq} &
		\sum_{k=1}^K \sum_{i=1}^m \sum_{j=0}^{k-1} \alpha_i \sigma_i^{k-j} \sqrt{\beta_i}\left\|u_i^{k+1}-u_i\right\|\left\|v_i^j-v_i^{j+1}\right\|_{H_{\beta_i}}+\sum_{k=1}^K \sum_{i=1}^m \alpha_i \sigma_i^k\left\|u_i^{k+1}-u_i\right\| \left\|e_i^0(u_i^1)\right\| \notag\\
		\leq & 
		\sum_{k=1}^K \sum_{i=1}^m \sum_{j=1}^{k-1} \alpha_i \sigma_i^{k-j} \sqrt{\beta_i}\left\|u_i^{k+1}-u_i\right\|\left\|v_i^j-v_i^{j+1}\right\|_{H_{\beta_i}}\notag\\
		&+\sum_{k=1}^K \sum_{i=1}^m \alpha_i \sigma_i^k\left\|u_i^{k+1}-u_i\right\|\left(\left\|e_i^0\left(u_i^1\right)\right\|+\sqrt{\beta_i}\left\|v_i^0-v_i^1\right\|_{H_{\beta_i}}\right) \\
		\leq & \sum_{k=1}^K \sum_{i=1}^m \sum_{j=1}^{k-1} \frac{\mu_i}{2} \alpha_i \sigma_i^{k-j}\left\|u_i^{k+1}-u_i\right\|^2+ \sum_{k=1}^K \sum_{i=1}^m \sum_{j=1}^{k-1}  \frac{1}{2 \mu_i}\alpha_i \sigma_i^{k-j}\beta_i\left\|v_i^j-v_i^{j+1}\right\|_{H_{\beta_i}}^2 \notag\\
		& +\sum_{k=1}^K \sum_{i=1}^m \frac{\mu_i}{2}  \alpha_i \sigma_i^k\left\|u_i^{k+1}-u_i\right\|^2+\sum_{k=1}^K \sum_{i=1}^m \frac{1}{2 \mu_i}  \alpha_i \sigma_i^k\left(\left\|e_i^0\left(u_i^1\right)\right\|+\sqrt{\beta_i}\left\|v_i^0-v_i^1\right\|_{H_{\beta_i}}\right)^2 \notag\\
		= & \sum_{k=1}^K \sum_{i=1}^m \sum_{j=0}^{k-1} \frac{\mu_i}{2}  \alpha_i \sigma_i^{k-j}\left\|u_i^{k+1}-u_i\right\|^2+ \sum_{k=1}^K \sum_{i=1}^m \sum_{j=1}^{k-1} \frac{1}{2 \mu_i} \alpha_i \sigma_i^{k-j} \beta_i\left\|v_i^j-v_i^{j+1}\right\|_{H_{\beta_i}}^2 \notag\\
		& +\sum_{k=1}^K \sum_{i=1}^m \frac{1}{2 \mu_i}  \alpha_i \sigma_i^k\left(\left\|e_i^0\left(u_i^1\right)\right\|+\sqrt{\beta_i}\left\|v_i^0-v_i^1\right\|_{H_{\beta_i}}\right)^2.\notag
	\end{align}
	It follows from \eqref{sigma} that $\sigma_i \in (0,1), \forall i \in [m]$. Then, by using the property of the geometric series, we can have
	\begin{align}
		& \sum_{k=1}^K\left(u^{k+1}-u\right)^{\top}\nabla_u L_{\alpha, \beta}\left(u^{k+1}, z^k, \lambda^k\right) \notag\\
		\leq & \sum_{k=1}^K \sum_{i=1}^m \alpha_i \frac{\mu_i}{2}  \frac{\sigma_i-\sigma_i^{k+1}}{1-\sigma_i}\left\|u_i^{k+1}-u_i\right\|^2+ \sum_{i=1}^m \sum_{j=1}^{K-1} \alpha_i \frac{1}{2 \mu_i} \frac{\sigma_i-\sigma_i^{K-j+1}}{1-\sigma_i} \beta_i\left\|v_i^j-v_i^{j+1}\right\|_{H_{\beta_i}}^2 \notag\\
		& +\sum_{i=1}^m \alpha_i \frac{1}{2 \mu_i} \frac{\sigma_i-\sigma_i^{K+1}}{1-\sigma_i}\left(\left\|e_i^0\left(u_i^1\right)\right\|+\sqrt{\beta_i}\left\|v_i^0-v_i^1\right\|_{H_{\beta_i}}\right)^2 \\
		\overset{\eqref{sigma}}{\leq} & \sum_{k=1}^K \sum_{i=1}^m \alpha_i \frac{\mu_i}{2} \frac{\sigma_i}{1-\sigma_i}\left\|u_i^{k+1}-u_i\right\|^2 + \sum_{k=1}^{K-1} \sum_{i=1}^m \alpha_i \frac{1}{2 \mu_i} \frac{\sigma_i}{1-\sigma_i} \beta_i\left\|v_i^k-v_i^{k+1}\right\|_{H_{\beta_i}}^2 \notag\\
		& +\sum_{i=1}^m \alpha_i \frac{1}{2 \mu_i} \frac{\sigma_i}{1-\sigma_i}\left(\left\|e_i^0\left(u_i^1\right)\right\|+\sqrt{\beta_i}\left\|v_i^0-v_i^1\right\|_{H_{\beta_i}}\right)^2, \quad \forall u \in \mathbb{R}^{m n}. \notag
	\end{align}
	This completes the proof.
\end{proof}

Now, with the help of \Cref{lemma1} and \Cref{lemma2}, we can show the convergence of our algorithm in the subsequent proof.
\begin{proof}[Proof of \Cref{thm_fedadmmIn}]
	First, from \eqref{F_of_w} and \eqref{strongly_convex}, we can have
	\begin{equation}\label{strongly_convex2}
		\begin{aligned}
			\left(w^{k+1}-w\right)^{\top} \left(F(w)-F(w^{k+1})\right) 
			\overset{\eqref{F_of_w}}{=}&
			\left(u^{k+1}-u\right)^{\top} I_\alpha \left(\nabla f(u)-\nabla f(u^{k+1})\right)\\ 
			=& 
			\sum_{i=1}^m \alpha_i\left(u_i^{k+1}-u_i\right)^{\top}\left(\nabla f_i\left(u_i\right)-\nabla f_i(u_i^{k+1})\right)\\
			\overset{\eqref{strongly_convex}}{\leq}& 
			-\sum_{i=1}^{m} \alpha_i c_i\left\|u_i-u_i^{k+1}\right\|^2 .
		\end{aligned}
	\end{equation}
	Then, using \eqref{res_lemma1} and \eqref{res_lemma2} established in \Cref{lemma1} and \Cref{lemma2}, respectively, we obtain
	\begin{align}\label{proof_the_1}
		& \sum_{k=1}^K \left(w^{k+1}-w\right)^{\top}F(w) \notag\\
		= & \sum_{k=1}^K \left(w^{k+1}-w\right)^{\top}F(w^{k+1})+\sum_{k=1}^K\left(w^{k+1}-w\right)^{\top}\left(F(w)-F(w^{k+1})\right)\notag\\
		\overset{\eqref{res_lemma1}, \eqref{strongly_convex2}}{\leq} & \sum_{k=1}^K\left(u^{k+1} - u\right)^{\top} \nabla_u L_{\alpha, \beta} \left(u^{k+1}, \lambda^k, z^k\right) + \frac{1}{2}\left(\left\|v^1-v\right\|_{H_1}^2-\left\|v^{K+1}-v\right\|_{H_1}^2\right)\notag\\
		&- \sum_{k=1}^K \frac{1}{2}\left\|v^k-v^{k+1}\right\|_{H_1}^2 -\sum_{k=1}^K\sum_{i=1}^{m} \alpha_i c_i\left\|u_i-u_i^{k+1}\right\|^2  \notag\\
		\overset{\eqref{res_lemma2}}{\leq} &
		\sum_{k=1}^K \sum_{i=1}^m \alpha_i\frac{\mu_i}{2} \frac{\sigma_i}{1-\sigma_i}\left\|u_i^{k+1}-u_i\right\|^2
		-
		\sum_{k=1}^K\sum_{i=1}^{m} \alpha_i c_i\left\|u_i-u_i^{k+1}\right\|^2\notag\\
		&+ \sum_{k=1}^{K-1} \sum_{i=1}^m \alpha_i\frac{1}{2 \mu_i} \frac{\sigma_i\beta_i}{1-\sigma_i} \left\|v_i^k-v_i^{k+1}\right\|_{{H_{\beta_i}}}^2 
		-
		\sum_{k=1}^K \sum_{i=1}^m \frac{\alpha_i}{2}\left\|v_i^k-v_i^{k+1}\right\|_{{H_{\beta_i}}}^2 \\
		& +\sum_{i=1}^m\alpha_i\frac{1}{2 \mu_i} \frac{\sigma_i}{1-\sigma_i}\left(\left\|e_i^0\left(u_i^1\right)\right\|+\sqrt{\beta_i}\left\|v_i^0-v_i^1\right\|_{H_{\beta_i}}\right)^2\notag\\
		& + \frac{1}{2}\left(\left\|v^1-v\right\|_{H_1}^2-\left\|v^{K+1}-v\right\|_{H_1}^2\right)\notag\\
		= &
		\sum_{k=1}^K \sum_{i=1}^m \alpha_i \left(\frac{\mu_i}{2} \frac{\sigma_i}{1-\sigma_i}-c_i\right)\left\|u_i^{k+1}-u_i\right\|^2 + \sum_{k=1}^{K-1} \sum_{i=1}^m \frac{\alpha_i}{2} \left(\frac{1}{\mu_i} \frac{\sigma_i\beta_i}{1-\sigma_i} - 1\right)\left\|v_i^k-v_i^{k+1}\right\|_{H_{\beta_i}}^2\notag\\
		&+\sum_{i=1}^m \alpha_i\frac{1}{2 \mu_i} \frac{\sigma_i}{1-\sigma_i}\left(\left\|e_i^0\left(u_i^1\right)\right\|+\sqrt{\beta_i}\left\|v_i^0-v_i^1\right\|_{H_{\beta_i}}\right)^2\notag\\
		& + \frac{1}{2}\left(\left\|v^1-v\right\|_{H_1}^2-\left\|v^{K+1}-v\right\|_{H_1}^2 - \left\|v^K-v^{K+1}\right\|_{H_1}^2\right). \notag
	\end{align}
	For the solution point $w^*$, it follows from \eqref{VI_opt} that $ F(w^{*}) = 0$.
	Setting $w=w^*$ in \eqref{proof_the_1}, together with the above property, for any integer $K>1$, we have
	\begin{align}\label{key_lemma2}
		&\sum_{k=1}^K \sum_{i=1}^m \alpha_i \left(c_i - \frac{\mu_i}{2} \frac{\sigma_i}{1-\sigma_i}\right)\left\|u_i^{k+1}-u_i^*\right\|^2  
		+ \sum_{k=1}^{K-1} \sum_{i=1}^m \frac{\alpha_i}{2} \left(1 - \frac{1}{\mu_i} \frac{\sigma_i\beta_i}{1-\sigma_i}\right)\left\|v_i^k-v_i^{k+1}\right\|_{H_{\beta_i}}^2 \notag\\
		\leq & \sum_{i=1}^m \alpha_i\frac{1}{2 \mu_i} \frac{\sigma_i}{1-\sigma_i}\left(\left\|e_i^0\left(u_i^1\right)\right\|+\sqrt{\beta_i}\left\|v_i^0-v_i^1\right\|_{H_{\beta_i}}\right)^2\\
		& + \frac{1}{2}\left(\left\|v^1-v^*\right\|_{H_1}^2-\left\|v^{K+1}-v^*\right\|_{H_1}^2 - \left\|v^K-v^{K+1}\right\|_{H_1}^2\right).\notag
	\end{align}
	Note that $\alpha_i>0, \forall i \in [m]$, and it follows from \eqref{mu_positive} that
	\begin{equation}
		\left(c_i - \frac{\mu_i}{2} \frac{\sigma_i}{1-\sigma_i}\right)>0 \quad\text { and }\quad \left(1 - \frac{1}{\mu_i} \frac{\sigma_i\beta_i}{1-\sigma_i}\right)>0, \quad \forall i \in [m].
	\end{equation}
	Therefore, the inequality \eqref{key_lemma2} implies that
	\begin{equation}\label{conv_u}
		\left\|u_i^{k+1}-u_i^*\right\| \stackrel{k \rightarrow \infty}{\longrightarrow} 0, \quad \left\|v_i^{k+1}-v_i^k\right\|_{H_{\beta_i}} \stackrel{k \rightarrow \infty}{\longrightarrow} 0, \quad \forall i \in [m].
	\end{equation}
	Using $\|v_i^{k+1}-v_i^k\|_{H_{\beta_i}} \stackrel{k \rightarrow \infty}{\longrightarrow} 0$ and \eqref{H_beta_i}, we can infer that
	\begin{equation}
		\left\|z^{k+1}-z^k\right\| \stackrel{k \rightarrow \infty}{\longrightarrow} 0, \quad \left\|\lambda_i^{k+1}-\lambda_i^k\right\| \stackrel{k \rightarrow \infty}{\longrightarrow} 0, \quad \forall i \in [m].
	\end{equation}
	Since $\|u_i^{k+1}-z^{k+1}\|=\frac{1}{\beta_i}\|\lambda_i^{k+1}-\lambda_i^k\|$, we can also have 
	\begin{equation}\label{xxx}
		\left\|u_i^{k+1}-z^{k+1}\right\| \stackrel{k \rightarrow \infty}{\longrightarrow} 0, \quad \forall i \in [m]
	\end{equation}
	Combining the facts that $u_i^k \stackrel{k \rightarrow \infty}{\longrightarrow} u_i^*$ in \eqref{conv_u}, $u_i^{*}=z^{*}$, and $\|u_i^{k+1}-z^{k+1}\| \stackrel{k \rightarrow \infty}{\longrightarrow} 0$ in \eqref{xxx}, one has 
	\begin{equation}
		z^k \stackrel{k \rightarrow \infty}{\longrightarrow} z^*.
	\end{equation}
	It follows from \eqref{conv_u} that for any $\epsilon>0$, there exists $k_0$, such that for all $k \geq k_0$, we have ${\|v_i^{k+1}-v_i^k\|_{H_{\beta_i}}<\epsilon}$ and $\sigma_i^k<\epsilon$. Then, for all $k \geq k_0$, it follows from \eqref{eik2} that
	\begin{equation*}
		\begin{aligned}
			\left\|e_i^k(u_i^{k+1})\right\| \leq & \sum_{j=0}^{k-1} \sigma_i^{k-j} \sqrt{\beta_i}\|v_i^j-v_i^{j+1}\|_{H_{\beta_i}}+\sigma_i^k\left\|e_i^0\left(u_i^1\right)\right\|  \\
			= & \sum_{j=0}^{k_0-1} \sigma_i^{k-j} \sqrt{\beta_i}\|v_i^j-v_i^{j+1}\|_{H_{\beta_i}}+\sum_{j=k_0}^{k-1} \sigma_i^{k-j} \sqrt{\beta_i}\|v_i^j-v_i^{j+1}\|_{H_{\beta_i}}+\sigma_i^k\left\|e_i^0\left(u_i^1\right)\right\| \\
			\leq & 
			\left(\sqrt{\beta_i} \max_{0 \leq j \leq k_0-1} \|v_i^j-v_i^{j+1}\|_{H_{\beta_i}} \sum_{j=0}^{k_0-1} \sigma_i^{k-k_0-j}\right) \sigma_i^{k_0}
			+\epsilon \sqrt{\beta_i} \sum_{j=k_0}^{k-1} \sigma_i^{k-j}
			+\epsilon \left\|e_i^0\left(u_i^1\right)\right\| \\
			\leq & 
			\epsilon\left(\sqrt{\beta_i} \max _{0 \leq j \leq k_0-1}\|v_i^j-v_i^{j+1}\|_{H_{\beta_i}} \sum_{j=0}^{k_0-1} \sigma_i^{k-k_0-j}+\sqrt{\beta_i} \sum_{j=k_0}^{k-1} \sigma_i^{k-j}+\left\|e_i^0\left(u_i^1\right)\right\|\right),
		\end{aligned}
	\end{equation*}
	which implies that
	\begin{equation}
		\left\|e_i^k(u_i^{k+1})\right\| \stackrel{k \rightarrow\infty}{\longrightarrow} 0 , \quad \forall i \in [m].
	\end{equation}
	From \eqref{VI_opt} and \eqref{error_i_k}, we can have $\lambda_i^*= \nabla f_i(u_i^*)$ and 
	\begin{equation}
		\lambda_i^k=\nabla f_i(u_i^{k+1})+\beta_i(u_i^{k+1}-z^k)-e_i^k(u_i^{k+1}).
	\end{equation}
	Then, we have
	\begin{equation}
		\lambda_i^k-\lambda_i^* = \nabla f_i(u_i^{k+1})-\nabla f_i(u_i^*)+\beta_i(u_i^{k+1}-u_i^k)+\beta_i(u_i^k-z^k)-e_i^k(u_i^{k+1}) .
	\end{equation}
	Note that $u_i^k \stackrel{k \rightarrow \infty}{\longrightarrow} u_i^*, u_i^k-z^k \stackrel{k \rightarrow \infty}{\longrightarrow} 0$, $e_i^k\left(u^{k+1}\right) \stackrel{k \rightarrow \infty}{\longrightarrow} 0$, and the gradient of $f_i$ is Lipschitz continuous (see \Cref{ass_lipschitz}), we have
	\begin{equation}
		\lambda_i^k \stackrel{k \rightarrow \infty}{\longrightarrow} \lambda_i^*, \quad \forall i \in [m]. 
	\end{equation}
	We thus complete the proof.
\end{proof}

\bibliographystyle{elsarticle-harv} 
\bibliography{./references.bib}

\begin{thebibliography}{44}
\expandafter\ifx\csname natexlab\endcsname\relax\def\natexlab#1{#1}\fi
\providecommand{\url}[1]{\texttt{#1}}
\providecommand{\href}[2]{#2}
\providecommand{\path}[1]{#1}
\providecommand{\DOIprefix}{doi:}
\providecommand{\ArXivprefix}{arXiv:}
\providecommand{\URLprefix}{URL: }
\providecommand{\Pubmedprefix}{pmid:}
\providecommand{\doi}[1]{\href{http://dx.doi.org/#1}{\path{#1}}}
\providecommand{\Pubmed}[1]{\href{pmid:#1}{\path{#1}}}
\providecommand{\bibinfo}[2]{#2}
\ifx\xfnm\relax \def\xfnm[#1]{\unskip,\space#1}\fi
\bibitem[{Acar et~al.(2021)Acar, Zhao, Navarro, Mattina, Whatmough and Saligrama}]{acar2021federated}
\bibinfo{author}{Acar, D.A.E.}, \bibinfo{author}{Zhao, Y.}, \bibinfo{author}{Navarro, R.M.}, \bibinfo{author}{Mattina, M.}, \bibinfo{author}{Whatmough, P.N.}, \bibinfo{author}{Saligrama, V.}, \bibinfo{year}{2021}.
\newblock \bibinfo{title}{Federated {{Learning Based}} on {{Dynamic Regularization}}} \DOIprefix\doi{10.48550/arXiv.2111.04263}, \href{http://arxiv.org/abs/2111.04263}{{\tt arXiv:2111.04263}}.
\bibitem[{Awheda and Schwartz(2016)}]{awheda2016exponential}
\bibinfo{author}{Awheda, M.D.}, \bibinfo{author}{Schwartz, H.M.}, \bibinfo{year}{2016}.
\newblock \bibinfo{title}{Exponential moving average based multiagent reinforcement learning algorithms}.
\newblock \bibinfo{journal}{Artificial Intelligence Review} \bibinfo{volume}{45}, \bibinfo{pages}{299--332}.
\newblock \DOIprefix\doi{10.1007/s10462-015-9447-5}.
\bibitem[{Bonawitz et~al.(2019)Bonawitz, Eichner, Grieskamp, Huba, Ingerman, Ivanov, Kiddon, Kone{\v c}n{\`y}, Mazzocchi and McMahan}]{bonawitz2019federated}
\bibinfo{author}{Bonawitz, K.}, \bibinfo{author}{Eichner, H.}, \bibinfo{author}{Grieskamp, W.}, \bibinfo{author}{Huba, D.}, \bibinfo{author}{Ingerman, A.}, \bibinfo{author}{Ivanov, V.}, \bibinfo{author}{Kiddon, C.}, \bibinfo{author}{Kone{\v c}n{\`y}, J.}, \bibinfo{author}{Mazzocchi, S.}, \bibinfo{author}{McMahan, B.}, \bibinfo{year}{2019}.
\newblock \bibinfo{title}{Towards federated learning at scale: {{System}} design}.
\newblock \bibinfo{journal}{Proceedings of machine learning and systems} \bibinfo{volume}{1}, \bibinfo{pages}{374--388}.
\bibitem[{Boyd et~al.(2011)Boyd, Parikh, Chu, Peleato and Eckstein}]{boyd2011distributed}
\bibinfo{author}{Boyd, S.}, \bibinfo{author}{Parikh, N.}, \bibinfo{author}{Chu, E.}, \bibinfo{author}{Peleato, B.}, \bibinfo{author}{Eckstein, J.}, \bibinfo{year}{2011}.
\newblock \bibinfo{title}{Distributed optimization and statistical learning via the alternating direction method of multipliers}.
\newblock \bibinfo{journal}{Foundations and Trends{\textregistered} in Machine learning} \bibinfo{volume}{3}, \bibinfo{pages}{1--122}.
\newblock \DOIprefix\doi{10.1561/2200000016}.
\bibitem[{Cai et~al.(2021)Cai, Ravichandran, Maji, Fowlkes, Tu and Soatto}]{cai2021exponential}
\bibinfo{author}{Cai, Z.}, \bibinfo{author}{Ravichandran, A.}, \bibinfo{author}{Maji, S.}, \bibinfo{author}{Fowlkes, C.}, \bibinfo{author}{Tu, Z.}, \bibinfo{author}{Soatto, S.}, \bibinfo{year}{2021}.
\newblock \bibinfo{title}{Exponential {{Moving Average Normalization}} for {{Self-Supervised}} and {{Semi-Supervised Learning}}}, in: \bibinfo{booktitle}{Proceedings of the {{IEEE}}/{{CVF Conference}} on {{Computer Vision}} and {{Pattern Recognition}}}, pp. \bibinfo{pages}{194--203}.
\bibitem[{Dinh et~al.(2020)Dinh, Tran and Nguyen}]{dinh2020personalized}
\bibinfo{author}{Dinh, C.T.}, \bibinfo{author}{Tran, N.}, \bibinfo{author}{Nguyen, J.}, \bibinfo{year}{2020}.
\newblock \bibinfo{title}{Personalized {{Federated Learning}} with {{Moreau Envelopes}}}, in: \bibinfo{booktitle}{Advances in {{Neural Information Processing Systems}}}, \bibinfo{publisher}{Curran Associates, Inc.}. pp. \bibinfo{pages}{21394--21405}.
\bibitem[{Geng et~al.(2023)Geng, Mou, Li, Li, Beyan, Decker and Rong}]{geng2023improved}
\bibinfo{author}{Geng, J.}, \bibinfo{author}{Mou, Y.}, \bibinfo{author}{Li, Q.}, \bibinfo{author}{Li, F.}, \bibinfo{author}{Beyan, O.}, \bibinfo{author}{Decker, S.}, \bibinfo{author}{Rong, C.}, \bibinfo{year}{2023}.
\newblock \bibinfo{title}{Improved {{Gradient Inversion Attacks}} and {{Defenses}} in {{Federated Learning}}}.
\newblock \bibinfo{journal}{IEEE Transactions on Big Data} , \bibinfo{pages}{1--13}\DOIprefix\doi{10.1109/TBDATA.2023.3239116}.
\bibitem[{Glowinski(2014)}]{glowinski2014alternating}
\bibinfo{author}{Glowinski, R.}, \bibinfo{year}{2014}.
\newblock \bibinfo{title}{On {{Alternating Direction Methods}} of {{Multipliers}}: {{A Historical Perspective}}}, in: \bibinfo{editor}{Fitzgibbon, W.}, \bibinfo{editor}{Kuznetsov, Y.A.}, \bibinfo{editor}{Neittaanm{\"a}ki, P.}, \bibinfo{editor}{Pironneau, O.} (Eds.), \bibinfo{booktitle}{Modeling, {{Simulation}} and {{Optimization}} for {{Science}} and {{Technology}}}. \bibinfo{publisher}{Springer Netherlands}, \bibinfo{address}{Dordrecht}. volume~\bibinfo{volume}{34}, pp. \bibinfo{pages}{59--82}.
\newblock \DOIprefix\doi{10.1007/978-94-017-9054-3_4}.
\bibitem[{Glowinski and Marroco(1975)}]{glowinski1975approximation}
\bibinfo{author}{Glowinski, R.}, \bibinfo{author}{Marroco, A.}, \bibinfo{year}{1975}.
\newblock \bibinfo{title}{Sur l'approximation, par {\'e}l{\'e}ments finis d'ordre un, et la r{\'e}solution, par p{\'e}nalisation-dualit{\'e} d'une classe de probl{\`e}mes de {{Dirichlet}} non lin{\'e}aires}.
\newblock \bibinfo{journal}{Revue fran{\c c}aise d'automatique, informatique, recherche op{\'e}rationnelle. Analyse num{\'e}rique} \bibinfo{volume}{9}, \bibinfo{pages}{41--76}.
\newblock \DOIprefix\doi{10.1051/m2an/197509R200411}.
\bibitem[{Glowinski et~al.(2020)Glowinski, Song and Yuan}]{glowinski2020admm}
\bibinfo{author}{Glowinski, R.}, \bibinfo{author}{Song, Y.}, \bibinfo{author}{Yuan, X.}, \bibinfo{year}{2020}.
\newblock \bibinfo{title}{An {{ADMM}} numerical approach to linear parabolic state constrained optimal control problems}.
\newblock \bibinfo{journal}{Numerische Mathematik} \bibinfo{volume}{144}, \bibinfo{pages}{931--966}.
\newblock \DOIprefix\doi{10.1007/s00211-020-01104-4}.
\bibitem[{Glowinski et~al.(2022)Glowinski, Song, Yuan and Yue}]{glowinski2022application}
\bibinfo{author}{Glowinski, R.}, \bibinfo{author}{Song, Y.}, \bibinfo{author}{Yuan, X.}, \bibinfo{author}{Yue, H.}, \bibinfo{year}{2022}.
\newblock \bibinfo{title}{Application of the {{Alternating Direction Method}} of {{Multipliers}} to {{Control Constrained Parabolic Optimal Control Problems}} and {{Beyond}}}.
\newblock \bibinfo{journal}{Annals of Applied Mathematics} \bibinfo{volume}{38}.
\newblock \DOIprefix\doi{10.4208/aam.OA-2022-0004}.
\bibitem[{Goldstein et~al.(2015)Goldstein, Li and Yuan}]{goldstein2015adaptive}
\bibinfo{author}{Goldstein, T.}, \bibinfo{author}{Li, M.}, \bibinfo{author}{Yuan, X.}, \bibinfo{year}{2015}.
\newblock \bibinfo{title}{Adaptive {{Primal-Dual Splitting Methods}} for {{Statistical Learning}} and {{Image Processing}}}, in: \bibinfo{booktitle}{Advances in {{Neural Information Processing Systems}}}, \bibinfo{publisher}{Curran Associates, Inc.}
\bibitem[{Gong et~al.(2022)Gong, Li and Freris}]{gong2022fedadmm}
\bibinfo{author}{Gong, Y.}, \bibinfo{author}{Li, Y.}, \bibinfo{author}{Freris, N.M.}, \bibinfo{year}{2022}.
\newblock \bibinfo{title}{{{FedADMM}}: {{A Robust Federated Deep Learning Framework}} with {{Adaptivity}} to {{System Heterogeneity}}}, in: \bibinfo{booktitle}{2022 {{IEEE}} 38th {{International Conference}} on {{Data Engineering}} ({{ICDE}})}, pp. \bibinfo{pages}{2575--2587}.
\newblock \DOIprefix\doi{10.1109/ICDE53745.2022.00238}.
\bibitem[{He et~al.(2000)He, Yang and Wang}]{he2000alternating}
\bibinfo{author}{He, B.}, \bibinfo{author}{Yang, H.}, \bibinfo{author}{Wang, S.}, \bibinfo{year}{2000}.
\newblock \bibinfo{title}{Alternating {{Direction Method}} with {{Self-Adaptive Penalty Parameters}} for {{Monotone Variational Inequalities}}}.
\newblock \bibinfo{journal}{Journal of Optimization Theory and Applications} \bibinfo{volume}{106}, \bibinfo{pages}{337--356}.
\newblock \DOIprefix\doi{10.1023/A:1004603514434}.
\bibitem[{He and Yuan(2018)}]{he2018class}
\bibinfo{author}{He, B.}, \bibinfo{author}{Yuan, X.}, \bibinfo{year}{2018}.
\newblock \bibinfo{title}{A class of {{ADMM-based}} algorithms for three-block separable convex programming}.
\newblock \bibinfo{journal}{Computational Optimization and Applications} \bibinfo{volume}{70}, \bibinfo{pages}{791--826}.
\newblock \DOIprefix\doi{10.1007/s10589-018-9994-1}.
\bibitem[{He et~al.(2016)He, Zhang, Ren and Sun}]{he2016deep}
\bibinfo{author}{He, K.}, \bibinfo{author}{Zhang, X.}, \bibinfo{author}{Ren, S.}, \bibinfo{author}{Sun, J.}, \bibinfo{year}{2016}.
\newblock \bibinfo{title}{Deep residual learning for image recognition}, in: \bibinfo{booktitle}{Proceedings of the {{IEEE}} Conference on Computer Vision and Pattern Recognition}, pp. \bibinfo{pages}{770--778}.
\bibitem[{Hestenes(1969)}]{hestenes1969multiplier}
\bibinfo{author}{Hestenes, M.R.}, \bibinfo{year}{1969}.
\newblock \bibinfo{title}{Multiplier and gradient methods}.
\newblock \bibinfo{journal}{Journal of optimization theory and applications} \bibinfo{volume}{4}, \bibinfo{pages}{303--320}.
\newblock \DOIprefix\doi{10.1007/BF00927673}.
\bibitem[{Kairouz et~al.(2021)Kairouz, McMahan, Avent, Bellet, Bennis, Bhagoji, Bonawitz, Charles, Cormode, Cummings, D'Oliveira, Eichner, Rouayheb, Evans, Gardner, Garrett, Gasc{\'o}n, Ghazi, Gibbons, Gruteser, Harchaoui, He, He, Huo, Hutchinson, Hsu, Jaggi, Javidi, Joshi, Khodak, Konecn{\'y}, Korolova, Koushanfar, Koyejo, Lepoint, Liu, Mittal, Mohri, Nock, {\"O}zg{\"u}r, Pagh, Qi, Ramage, Raskar, Raykova, Song, Song, Stich, Sun, Suresh, Tram{\`e}r, Vepakomma, Wang, Xiong, Xu, Yang, Yu, Yu and Zhao}]{kairouz2021advancesa}
\bibinfo{author}{Kairouz, P.}, \bibinfo{author}{McMahan, H.B.}, \bibinfo{author}{Avent, B.}, \bibinfo{author}{Bellet, A.}, \bibinfo{author}{Bennis, M.}, \bibinfo{author}{Bhagoji, A.N.}, \bibinfo{author}{Bonawitz, K.}, \bibinfo{author}{Charles, Z.}, \bibinfo{author}{Cormode, G.}, \bibinfo{author}{Cummings, R.}, \bibinfo{author}{D'Oliveira, R.G.L.}, \bibinfo{author}{Eichner, H.}, \bibinfo{author}{Rouayheb, S.E.}, \bibinfo{author}{Evans, D.}, \bibinfo{author}{Gardner, J.}, \bibinfo{author}{Garrett, Z.}, \bibinfo{author}{Gasc{\'o}n, A.}, \bibinfo{author}{Ghazi, B.}, \bibinfo{author}{Gibbons, P.B.}, \bibinfo{author}{Gruteser, M.}, \bibinfo{author}{Harchaoui, Z.}, \bibinfo{author}{He, C.}, \bibinfo{author}{He, L.}, \bibinfo{author}{Huo, Z.}, \bibinfo{author}{Hutchinson, B.}, \bibinfo{author}{Hsu, J.}, \bibinfo{author}{Jaggi, M.}, \bibinfo{author}{Javidi, T.}, \bibinfo{author}{Joshi, G.}, \bibinfo{author}{Khodak, M.}, \bibinfo{author}{Konecn{\'y}, J.}, \bibinfo{author}{Korolova, A.}, \bibinfo{author}{Koushanfar, F.}, \bibinfo{author}{Koyejo, S.}, \bibinfo{author}{Lepoint, T.}, \bibinfo{author}{Liu, Y.}, \bibinfo{author}{Mittal, P.}, \bibinfo{author}{Mohri, M.}, \bibinfo{author}{Nock, R.}, \bibinfo{author}{{\"O}zg{\"u}r, A.}, \bibinfo{author}{Pagh, R.}, \bibinfo{author}{Qi, H.}, \bibinfo{author}{Ramage, D.}, \bibinfo{author}{Raskar, R.}, \bibinfo{author}{Raykova, M.}, \bibinfo{author}{Song, D.}, \bibinfo{author}{Song, W.}, \bibinfo{author}{Stich, S.U.}, \bibinfo{author}{Sun, Z.}, \bibinfo{author}{Suresh, A.T.}, \bibinfo{author}{Tram{\`e}r, F.}, \bibinfo{author}{Vepakomma, P.}, \bibinfo{author}{Wang, J.}, \bibinfo{author}{Xiong, L.}, \bibinfo{author}{Xu, Z.}, \bibinfo{author}{Yang, Q.}, \bibinfo{author}{Yu, F.X.}, \bibinfo{author}{Yu, H.}, \bibinfo{author}{Zhao, S.}, \bibinfo{year}{2021}.
\newblock \bibinfo{title}{Advances and {{Open Problems}} in {{Federated Learning}}}.
\newblock \bibinfo{journal}{Foundations and Trends{\textregistered} in Machine Learning} \bibinfo{volume}{14}, \bibinfo{pages}{1--210}.
\newblock \DOIprefix\doi{10.1561/2200000083}.
\bibitem[{Karimireddy et~al.(2020)Karimireddy, Kale, Mohri, Reddi, Stich and Suresh}]{karimireddy2020scaffold}
\bibinfo{author}{Karimireddy, S.P.}, \bibinfo{author}{Kale, S.}, \bibinfo{author}{Mohri, M.}, \bibinfo{author}{Reddi, S.}, \bibinfo{author}{Stich, S.}, \bibinfo{author}{Suresh, A.T.}, \bibinfo{year}{2020}.
\newblock \bibinfo{title}{{{SCAFFOLD}}: {{Stochastic Controlled Averaging}} for {{Federated Learning}}}, in: \bibinfo{booktitle}{Proceedings of the 37th {{International Conference}} on {{Machine Learning}}}, \bibinfo{publisher}{PMLR}. pp. \bibinfo{pages}{5132--5143}.
\bibitem[{Krizhevsky(2009)}]{krizhevsky2009learning}
\bibinfo{author}{Krizhevsky, A.}, \bibinfo{year}{2009}.
\newblock \bibinfo{title}{Learning Multiple Layers of Features from Tiny Images}.
\newblock \bibinfo{type}{Technical Report}. University of Toronto.
\bibitem[{LeCun et~al.(1998)LeCun, Bottou, Bengio and Haffner}]{lecun1998gradientbased}
\bibinfo{author}{LeCun, Y.}, \bibinfo{author}{Bottou, L.}, \bibinfo{author}{Bengio, Y.}, \bibinfo{author}{Haffner, P.}, \bibinfo{year}{1998}.
\newblock \bibinfo{title}{Gradient-based learning applied to document recognition}.
\newblock \bibinfo{journal}{Proceedings of the IEEE} \bibinfo{volume}{86}, \bibinfo{pages}{2278--2324}.
\newblock \DOIprefix\doi{10.1109/5.726791}.
\bibitem[{Li et~al.(2021)Li, Wen, Wu, Hu, Wang, Li, Liu and He}]{li2021survey}
\bibinfo{author}{Li, Q.}, \bibinfo{author}{Wen, Z.}, \bibinfo{author}{Wu, Z.}, \bibinfo{author}{Hu, S.}, \bibinfo{author}{Wang, N.}, \bibinfo{author}{Li, Y.}, \bibinfo{author}{Liu, X.}, \bibinfo{author}{He, B.}, \bibinfo{year}{2021}.
\newblock \bibinfo{title}{A {{Survey}} on {{Federated Learning Systems}}: {{Vision}}, {{Hype}} and {{Reality}} for {{Data Privacy}} and {{Protection}}}.
\newblock \bibinfo{journal}{IEEE Transactions on Knowledge and Data Engineering} , \bibinfo{pages}{1--1}\DOIprefix\doi{10.1109/TKDE.2021.3124599}.
\bibitem[{Li et~al.(2020a)Li, Sahu, Talwalkar and Smith}]{li2020federated}
\bibinfo{author}{Li, T.}, \bibinfo{author}{Sahu, A.K.}, \bibinfo{author}{Talwalkar, A.}, \bibinfo{author}{Smith, V.}, \bibinfo{year}{2020}a.
\newblock \bibinfo{title}{Federated {{Learning}}: {{Challenges}}, {{Methods}}, and {{Future Directions}}}.
\newblock \bibinfo{journal}{IEEE Signal Processing Magazine} \bibinfo{volume}{37}, \bibinfo{pages}{50--60}.
\newblock \DOIprefix\doi{10.1109/MSP.2020.2975749}.
\bibitem[{Li et~al.(2020b)Li, Sahu, Zaheer, Sanjabi, Talwalkar and Smith}]{li2020federateda}
\bibinfo{author}{Li, T.}, \bibinfo{author}{Sahu, A.K.}, \bibinfo{author}{Zaheer, M.}, \bibinfo{author}{Sanjabi, M.}, \bibinfo{author}{Talwalkar, A.}, \bibinfo{author}{Smith, V.}, \bibinfo{year}{2020}b.
\newblock \bibinfo{title}{Federated {{Optimization}} in {{Heterogeneous Networks}}}.
\newblock \bibinfo{journal}{Proceedings of Machine Learning and Systems} \bibinfo{volume}{2}, \bibinfo{pages}{429--450}.
\bibitem[{Li et~al.(2019)Li, Huang, Yang, Wang and Zhang}]{li2019convergence}
\bibinfo{author}{Li, X.}, \bibinfo{author}{Huang, K.}, \bibinfo{author}{Yang, W.}, \bibinfo{author}{Wang, S.}, \bibinfo{author}{Zhang, Z.}, \bibinfo{year}{2019}.
\newblock \bibinfo{title}{On the {{Convergence}} of {{FedAvg}} on {{Non-IID Data}}}, in: \bibinfo{booktitle}{International {{Conference}} on {{Learning Representations}}}.
\bibitem[{Liu and Nocedal(1989)}]{liu1989limited}
\bibinfo{author}{Liu, D.C.}, \bibinfo{author}{Nocedal, J.}, \bibinfo{year}{1989}.
\newblock \bibinfo{title}{On the limited memory {{BFGS}} method for large scale optimization}.
\newblock \bibinfo{journal}{Mathematical Programming} \bibinfo{volume}{45}, \bibinfo{pages}{503--528}.
\newblock \DOIprefix\doi{10.1007/BF01589116}.
\bibitem[{McMahan et~al.(2017)McMahan, Moore, Ramage, Hampson and y~Arcas}]{mcmahan2017communicationefficient}
\bibinfo{author}{McMahan, B.}, \bibinfo{author}{Moore, E.}, \bibinfo{author}{Ramage, D.}, \bibinfo{author}{Hampson, S.}, \bibinfo{author}{y~Arcas, B.A.}, \bibinfo{year}{2017}.
\newblock \bibinfo{title}{Communication-{{Efficient Learning}} of {{Deep Networks}} from {{Decentralized Data}}}, in: \bibinfo{booktitle}{Proceedings of the 20th {{International Conference}} on {{Artificial Intelligence}} and {{Statistics}}}, \bibinfo{publisher}{PMLR}. pp. \bibinfo{pages}{1273--1282}.
\bibitem[{Powell(1969)}]{powell1969method}
\bibinfo{author}{Powell, M.J.}, \bibinfo{year}{1969}.
\newblock \bibinfo{title}{A method for nonlinear constraints in minimization problems}.
\newblock \bibinfo{journal}{Optimization} , \bibinfo{pages}{283--298}.
\bibitem[{Song et~al.(2016)Song, Yoon and Pavlovic}]{song2016fast}
\bibinfo{author}{Song, C.}, \bibinfo{author}{Yoon, S.}, \bibinfo{author}{Pavlovic, V.}, \bibinfo{year}{2016}.
\newblock \bibinfo{title}{Fast {{ADMM Algorithm}} for {{Distributed Optimization}} with {{Adaptive Penalty}}}.
\newblock \bibinfo{journal}{Proceedings of the AAAI Conference on Artificial Intelligence} \bibinfo{volume}{30}.
\newblock \DOIprefix\doi{10.1609/aaai.v30i1.10069}.
\bibitem[{Song et~al.(2023)Song, Yuan and Yue}]{song2023admmpinnsa}
\bibinfo{author}{Song, Y.}, \bibinfo{author}{Yuan, X.}, \bibinfo{author}{Yue, H.}, \bibinfo{year}{2023}.
\newblock \bibinfo{title}{The {{ADMM-PINNs Algorithmic Framework}} for {{Nonsmooth PDE-Constrained Optimization}}: {{A Deep Learning Approach}}}.
\newblock \bibinfo{journal}{arXiv preprint arXiv:2302.08309} \DOIprefix\doi{10.48550/arXiv.2302.08309}, \href{http://arxiv.org/abs/2302.08309}{{\tt arXiv:2302.08309}}.
\bibitem[{Tan et~al.(2023)Tan, Yu, Cui and Yang}]{tan2023personalized}
\bibinfo{author}{Tan, A.Z.}, \bibinfo{author}{Yu, H.}, \bibinfo{author}{Cui, L.}, \bibinfo{author}{Yang, Q.}, \bibinfo{year}{2023}.
\newblock \bibinfo{title}{Towards {{Personalized Federated Learning}}}.
\newblock \bibinfo{journal}{IEEE Transactions on Neural Networks and Learning Systems} \bibinfo{volume}{34}, \bibinfo{pages}{9587--9603}.
\newblock \DOIprefix\doi{10.1109/TNNLS.2022.3160699}.
\bibitem[{Wang et~al.(2022)Wang, Marella and Anderson}]{wang2022fedadmm}
\bibinfo{author}{Wang, H.}, \bibinfo{author}{Marella, S.}, \bibinfo{author}{Anderson, J.}, \bibinfo{year}{2022}.
\newblock \bibinfo{title}{{{FedADMM}}: {{A}} federated primal-dual algorithm allowing partial participation}, in: \bibinfo{booktitle}{2022 {{IEEE}} 61st {{Conference}} on {{Decision}} and {{Control}} ({{CDC}})}, pp. \bibinfo{pages}{287--294}.
\newblock \DOIprefix\doi{10.1109/CDC51059.2022.9992745}.
\bibitem[{Wang et~al.(2020)Wang, Liu, Liang, Joshi and Poor}]{wang2020tackling}
\bibinfo{author}{Wang, J.}, \bibinfo{author}{Liu, Q.}, \bibinfo{author}{Liang, H.}, \bibinfo{author}{Joshi, G.}, \bibinfo{author}{Poor, H.V.}, \bibinfo{year}{2020}.
\newblock \bibinfo{title}{Tackling the objective inconsistency problem in heterogeneous federated optimization}, in: \bibinfo{booktitle}{Proceedings of the 34th {{International Conference}} on {{Neural Information Processing Systems}}}, \bibinfo{publisher}{Curran Associates Inc.}, \bibinfo{address}{Red Hook, NY, USA}. pp. \bibinfo{pages}{7611--7623}.
\bibitem[{Wang et~al.(2023a)Wang, Xu, Wang, Chang, Quek and Sun}]{wang2023admm}
\bibinfo{author}{Wang, S.}, \bibinfo{author}{Xu, Y.}, \bibinfo{author}{Wang, Z.}, \bibinfo{author}{Chang, T.H.}, \bibinfo{author}{Quek, T.Q.S.}, \bibinfo{author}{Sun, D.}, \bibinfo{year}{2023}a.
\newblock \bibinfo{title}{Beyond {{ADMM}}: {{A Unified Client-Variance-Reduced Adaptive Federated Learning Framework}}}.
\newblock \bibinfo{journal}{Proceedings of the AAAI Conference on Artificial Intelligence} \bibinfo{volume}{37}, \bibinfo{pages}{10175--10183}.
\newblock \DOIprefix\doi{10.1609/aaai.v37i8.26212}.
\bibitem[{Wang et~al.(2023b)Wang, Song and Zuazua}]{wang2023approximatea}
\bibinfo{author}{Wang, Z.}, \bibinfo{author}{Song, Y.}, \bibinfo{author}{Zuazua, E.}, \bibinfo{year}{2023}b.
\newblock \bibinfo{title}{Approximate and weighted data reconstruction attack in federated learning}.
\newblock \bibinfo{journal}{arXiv preprint arXiv:2308.06822} \DOIprefix\doi{10.48550/arXiv.2308.06822}, \href{http://arxiv.org/abs/2308.06822}{{\tt arXiv:2308.06822}}.
\bibitem[{Xiao et~al.(2024)Xiao, Li and Li}]{xiao2024privacypreserving}
\bibinfo{author}{Xiao, D.}, \bibinfo{author}{Li, J.}, \bibinfo{author}{Li, M.}, \bibinfo{year}{2024}.
\newblock \bibinfo{title}{Privacy-{{Preserving Federated Compressed Learning Against Data Reconstruction Attacks Based}} on~{{Secure Data}}}, in: \bibinfo{editor}{Luo, B.}, \bibinfo{editor}{Cheng, L.}, \bibinfo{editor}{Wu, Z.G.}, \bibinfo{editor}{Li, H.}, \bibinfo{editor}{Li, C.} (Eds.), \bibinfo{booktitle}{Neural {{Information Processing}}}, \bibinfo{publisher}{Springer Nature}, \bibinfo{address}{Singapore}. pp. \bibinfo{pages}{325--339}.
\newblock \DOIprefix\doi{10.1007/978-981-99-8184-7_25}.
\bibitem[{Xu et~al.(2017)Xu, Liu, Lin and Yang}]{xu2017admm}
\bibinfo{author}{Xu, Y.}, \bibinfo{author}{Liu, M.}, \bibinfo{author}{Lin, Q.}, \bibinfo{author}{Yang, T.}, \bibinfo{year}{2017}.
\newblock \bibinfo{title}{{{ADMM}} without a {{Fixed Penalty Parameter}}: {{Faster Convergence}} with {{New Adaptive Penalization}}}, in: \bibinfo{booktitle}{Advances in {{Neural Information Processing Systems}}}, \bibinfo{publisher}{Curran Associates, Inc.}
\bibitem[{Yang et~al.(2023)Yang, Ma, Xiao, Liu, Li and Zhang}]{yang2023reveal}
\bibinfo{author}{Yang, Y.}, \bibinfo{author}{Ma, Z.}, \bibinfo{author}{Xiao, B.}, \bibinfo{author}{Liu, Y.}, \bibinfo{author}{Li, T.}, \bibinfo{author}{Zhang, J.}, \bibinfo{year}{2023}.
\newblock \bibinfo{title}{Reveal {{Your Images}}: {{Gradient Leakage Attack}} against {{Unbiased Sampling-Based Secure Aggregation}}}.
\newblock \bibinfo{journal}{IEEE Transactions on Knowledge and Data Engineering} , \bibinfo{pages}{1--14}\DOIprefix\doi{10.1109/TKDE.2023.3271432}.
\bibitem[{Yue et~al.(2018)Yue, Yang, Wang and Yuan}]{yue2018implementing}
\bibinfo{author}{Yue, H.}, \bibinfo{author}{Yang, Q.}, \bibinfo{author}{Wang, X.}, \bibinfo{author}{Yuan, X.}, \bibinfo{year}{2018}.
\newblock \bibinfo{title}{Implementing the {{Alternating Direction Method}} of {{Multipliers}} for {{Big Datasets}}: {{A Case Study}} of {{Least Absolute Shrinkage}} and {{Selection Operator}}}.
\newblock \bibinfo{journal}{SIAM Journal on Scientific Computing} \bibinfo{volume}{40}, \bibinfo{pages}{A3121--A3156}.
\newblock \DOIprefix\doi{10.1137/17M1146567}.
\bibitem[{Zhang et~al.(2017)Zhang, Li, Song and Wang}]{zhang2017alternating}
\bibinfo{author}{Zhang, K.}, \bibinfo{author}{Li, J.}, \bibinfo{author}{Song, Y.}, \bibinfo{author}{Wang, X.}, \bibinfo{year}{2017}.
\newblock \bibinfo{title}{An alternating direction method of multipliers for elliptic equation constrained optimization problem}.
\newblock \bibinfo{journal}{Science China Mathematics} \bibinfo{volume}{60}, \bibinfo{pages}{361--378}.
\newblock \DOIprefix\doi{10.1007/s11425-015-0522-3}.
\bibitem[{Zhang et~al.(2022)Zhang, Gao, He, Zhang, Krishnamachari and Avestimehr}]{zhang2022federated}
\bibinfo{author}{Zhang, T.}, \bibinfo{author}{Gao, L.}, \bibinfo{author}{He, C.}, \bibinfo{author}{Zhang, M.}, \bibinfo{author}{Krishnamachari, B.}, \bibinfo{author}{Avestimehr, A.S.}, \bibinfo{year}{2022}.
\newblock \bibinfo{title}{Federated {{Learning}} for the {{Internet}} of {{Things}}: {{Applications}}, {{Challenges}}, and {{Opportunities}}}.
\newblock \bibinfo{journal}{IEEE Internet of Things Magazine} \bibinfo{volume}{5}, \bibinfo{pages}{24--29}.
\newblock \DOIprefix\doi{10.1109/IOTM.004.2100182}.
\bibitem[{Zhang et~al.(2021)Zhang, Hong, Dhople, Yin and Liu}]{zhang2021fedpd}
\bibinfo{author}{Zhang, X.}, \bibinfo{author}{Hong, M.}, \bibinfo{author}{Dhople, S.}, \bibinfo{author}{Yin, W.}, \bibinfo{author}{Liu, Y.}, \bibinfo{year}{2021}.
\newblock \bibinfo{title}{{{FedPD}}: {{A Federated Learning Framework With Adaptivity}} to {{Non-IID Data}}}.
\newblock \bibinfo{journal}{IEEE Transactions on Signal Processing} \bibinfo{volume}{69}, \bibinfo{pages}{6055--6070}.
\newblock \DOIprefix\doi{10.1109/TSP.2021.3115952}.
\bibitem[{Zhou and Li(2023)}]{zhou2023federated}
\bibinfo{author}{Zhou, S.}, \bibinfo{author}{Li, G.Y.}, \bibinfo{year}{2023}.
\newblock \bibinfo{title}{Federated {{Learning Via Inexact ADMM}}}.
\newblock \bibinfo{journal}{IEEE Transactions on Pattern Analysis and Machine Intelligence} \bibinfo{volume}{45}, \bibinfo{pages}{9699--9708}.
\newblock \DOIprefix\doi{10.1109/TPAMI.2023.3243080}.
\bibitem[{Zinkevich et~al.(2010)Zinkevich, Weimer, Li and Smola}]{zinkevich2010parallelized}
\bibinfo{author}{Zinkevich, M.}, \bibinfo{author}{Weimer, M.}, \bibinfo{author}{Li, L.}, \bibinfo{author}{Smola, A.}, \bibinfo{year}{2010}.
\newblock \bibinfo{title}{Parallelized {{Stochastic Gradient Descent}}}, in: \bibinfo{booktitle}{Advances in {{Neural Information Processing Systems}}}, \bibinfo{publisher}{Curran Associates, Inc.}

\end{thebibliography}

\end{document}